\documentclass{article}

\usepackage{arxiv}

\usepackage[utf8]{inputenc} 
\usepackage[T1]{fontenc} 
\usepackage{hyperref}  
\usepackage{url}   
\usepackage{booktabs}  
\usepackage{amsfonts}  
\usepackage{nicefrac}  
\usepackage{microtype}  
\usepackage{graphicx}
\usepackage{caption}
\usepackage{subcaption}
\usepackage{xspace}
\usepackage{geometry}
\usepackage{amsmath,amssymb,amsfonts,dsfont}
\usepackage{breqn}
\usepackage{lmodern}
\usepackage[lined,  ruled, linesnumbered]{algorithm2e}
\usepackage{xcolor}

\title{Learning Enhanced Optimisation for Routing Problems}
\author{Nasrin Sultana, Jeffrey Chan, Tabinda Sarwar, Babak Abbasi, A. K. Qin }

\begin{document}
\maketitle

\begin{abstract}

Deep learning approaches have shown promising results in solving routing problems. However, there is still a substantial gap in solution quality between machine learning and operations research algorithms. Recently, another line of research has been introduced that fuses the strengths of machine learning and operational research algorithms. In particular, search perturbation operators have been used to improve the solution. Nevertheless, using the perturbation may not guarantee a quality solution. This paper presents "Learning to Guide Local Search" (L2GLS), a learning-based approach for routing problems that uses a penalty term and reinforcement learning to adaptively adjust search efforts. L2GLS combines local search (LS) operators' strengths with penalty terms to escape local optimals. Routing problems have many practical applications, often presetting larger instances that are still challenging for many existing algorithms introduced in the learning to optimise field. We show that L2GLS achieves the new state-of-the-art results on larger TSP and CVRP over other machine learning methods.

\end{abstract}


\section{Introduction}

Routing problems are an important class of combinatorial optimisation problems (COPs), which have many real-life applications, e.g., supply chain and warehouse management, aviation planning, healthcare scheduling, and hardware design~\cite{baldacci2010some}. Among the routing problems, the Travelling Salesman Problem (TSP) and Vehicle Routing Problem (VRP) are the two most prevalent ones. TSP is defined as finding a tour of all cities/locations where each city is visited only once and has a minimum total distance travelled. The VRP is to find the optimal set of routes for a fleet of vehicles. VRP has many variants; one example is Capacitated vehicle routing problems (CVRP) that aims to find a set of routes with minimum cost to fulfil customers' demands without violating the vehicle capacity constraints. 

Routing problems~\cite{dantzig1954solution} can be solved efficiently with many optimisation methods, including Branch-and-Bound~\cite{held1962dynamic}, Local Search (LS)~\cite{croes1958method}, Lagrangian Relaxation~\cite{fisher1981lagrangian} and Tabu Search~\cite{glover1993user}. A highly specialised algorithm was designed for TSP, namely, Concorde~\cite{applegate2006concorde}, which is widely used as an exact TSP solver for large instances. Nevertheless, solving larger instances is difficult for Concorde because of the exponential growth in execution time with increasing instance size. Another approach is manually designed heuristics that can discover close-to-optimal results~\cite{helsgaun2017extension}~\cite{lin1973effective}. For example for CVRP, LKH3~\cite{helsgaun2017extension} is widely used and is a penalty-function-based extension of the classical Lin-Kernighan heuristic~\cite{helsgaun2017extension}. These two-classic algorithms \cite{applegate2006concorde}\cite{helsgaun2017extension} are able to provide a solution with an average cost of 7.77 for 100 cities TSP and 15.65 for 100 customers CVRP  but with a long-running time. For instance, for 100 nodes TSP instances, Concorde~\cite{applegate2006traveling} needs on average one hour to solve one instance, and for 100 nodes CVRP instances, LKH3 needs on average 13 hours to solve an instance. 

As highlighted, the traditional solvers have many issues. Recently, Deep Learning (DL) is used to learn heuristics for solving routing problems~\cite{kool2018attention}~\cite{bello2016neural}~\cite{nazari2018reinforcement} automatically. They typically develop DL and Reinforcement Learning (RL) frameworks to predict a complete solution from scratch. Their frameworks have shown their competency to obtain an approximate solution to solve RPs. Once the learning-based methods are trained, the execution time is fast. However, current learning-based methods have shortcomings. First, these learning-based methods are designed to act as a constructive heuristic, such as constructing a solution to the TSP sequence one node at a time. However, this model might not be scalable when there are a larger number of nodes (customers). Second, learning-based methods rely on gradient information to guide the search (greedy search or beam search), which may not be available because the solution space is undifferentiable or finding a differentiable surrogate is difficult. In addition, another issue is that most of the work in the Learning to Optimise (L2O) field evaluated their models on randomly generated and easy to solve TSP and CVRP instances and focused on solving instances up to 100 nodes. Scaling to large and real-world instances is still an open question. Solving larger problems is highly time-consuming for most of the current operation research algorithms~\cite{helsgaun2017extension}~\cite{lai2016tabu}, and~\cite{wei2018simulated}. Therefore, all approaches including exact, heuristics and learning-based, have strengths and weaknesses. Naturally, one may want to combine the strengths of operation research and learning-based algorithms to build a mechanism for solving large routing problems efficiently.

There have been some previous approaches based on operation research algorithms and learning-based heuristics and tested on TSP~\cite{wu2019learning} and CVRP~\cite{lu2019learning}~\cite{chen2019learning}~\cite{hottung2019neural}. Hottung et al.~\cite{hottung2019neural} used destroy operators to improve solutions, while L2I~\cite{lu2019learning} used perturbation to escape local minima and combined with search operators~\cite{lu2019learning} to improve solutions. However, perturbation operators have some issues, such as they do not guarantee an improvement in the quality of the returned solution~\cite{misztal2019impact} for all the problems. L2I~\cite{lu2019learning} demonstrated that when the magnitude of perturbation is too large, the resulting solution generally becomes much worse, and the algorithm will take many improvement steps to repair the deterioration. 

To address this challenge, we design a method that can scale to large instances that are quite common in real-world problems. In our work, instead of using perturbation operators to avoid local minimum, our method augments the cost function to include a set of penalty terms~\cite{voudouris1999guided} and passes the new modification to LS. The goal is to escape the local minimum and guide the LS process to distribute search efforts.
Furthermore, previous works never investigated which combination of search operators was successful for various routing problems in the L2O field. The current literature focuses on algorithm optimisation performance rather than a thorough examination of various operators. Since selected operators have an impact on the algorithm's efficiency, we also need to analyse them thoroughly. Hence, we analysed many combinations of LS operators and constructed a rich set of LS operators to improve the solution~\cite{sengupta2019local}~\cite{lourencco2019iterated}~\cite{gendreau1992new}. One of the important observations is that successful heuristics for the route optimisation problems do not necessarily require new operators~\cite{arnold2019knowledge}. We investigate multiple operators and their combinations for the case of LS here. We consider a group of operators, namely 2-opt~\cite{croes1958method}, relocate~\cite{gendreau1992new}, three permutations~\cite{sengupta2019local} and location swap~\cite{resende2007fast} (all operators are described in Section~\ref{sec:local}). The idea behind our method is to find a better solution through search operations by taking improvement progressively into account, learning using RL instead of handcrafted heuristics. 

Our proposed approach can improve the initial solution, progressively guiding the search process using penalty terms among feasible solutions using LS operator, followed by a reinforcement-based manager to learn the set of LS operators that are more effective for the RP. 
{Our main contributions are as follows:}
\begin{itemize}

\item We propose a scalable learning-based algorithm 
that solve large scale routing problems, achieving new state-of-the-art results. Proposed model can generalise on the benchmark datasets, such as TSPLIB~\cite{reinelt1991tsplib} and for CVRP, mentioned in Uchoa et al.~\cite{uchoa2017new}. Our algorithm outperforms Concorde on TSP in terms of computation time and solution quality for randomly generated instances, that has gained attention in the last few years for L2O algorithms.
\item We propose a framework, where LS operators place in action. LS uses a set of penalty terms when stuck in local minima and searches the promising areas of the search space, i.e., instead of changing the search direction, our method dynamically changing the objective.
\item The recent line of research using LS operators shows the potential to solve routing problems using RL. Nevertheless, they used search operators without investigating which combination of operators are effective. We studied a set of LS operators and a learning approach that automatically finds an effective set of operators and the order to apply for routing problems. 
\end{itemize}
\section{Related Work}\label{sec:related}
DL and RL have recently been proposed to solve COPs. Learning-based method for routing problems can be categorised as constructive~\cite{vinyals2015pointer}~\cite{bello2016neural}~\cite{nazari2018reinforcement}~\cite{deudon2018learning}~\cite{kool2018attention}~\cite{sultana2020learning}~\cite{sultana2020learningb}~\cite{kwon2020pomo}~\cite{bresson2021transformer} and improvement heuristics~\cite{chen2019learning}~\cite{hottung2019neural}~\cite{da2020learning}~\cite{lu2019learning}~\cite{wu2019learning}~\cite{gao2020learn}. Constructive methods tend to predict the next node given a partially constructed tour and build up to a solution node by node. Improvement heuristics improve a solution by iteratively performing search based on LS heuristics to improve the solution~\cite{li2016learning}, where a neural network guides a LS algorithm that iteratively finds promising solutions.

A pioneering work in this area is Pointer Networks (PN)~\cite{vinyals2015pointer} based on attention mechanism and a supervised learning approach to solve TSP. It relies on having an existing set of good solutions or a good solver to generate training instances. One recent paper used a supervised learning approach to train a small-scale model, which could be repetitively used to build heat maps for TSP instances of arbitrarily large instances based on graph sampling, graph converting and heat maps merging techniques. Then the heat maps are fed into a reinforcement learning approach to guide the search for high-quality solutions based on the Monte Carlo tree search~\cite{fu2020generalize}. In~\cite{nazari2018reinforcement}, a constructive neural method was proposed that uses a recurrent neural network (RNN) decoder and an attention mechanism to build solutions for the CVRP and uses an actor-critic RL for training. A graph attention network is used in the method called AM ~\cite{kool2018attention} and generates solutions for different routing problems trained via RL, including TSP and CVRP. They trained their model using policy gradient RL with a baseline based on a deterministic greedy roll-out. ERRL~\cite{sultana2020learningb} is another RL based approach that introduces more exploration via stochastic policies. POMO~\cite{kwon2020pomo} introduced an end-to-end approach for building a heuristic solver based on policy optimisation with multiple optima. Bresson et al.~\cite{bresson2021transformer} proposed adapting the popular Transformer architecture to solve TSP. Training is carried out through RL (without TSP training solutions) and decoding uses beam search. Other lines of research are Deep Policy Dynamic Programming that aims to combine the strengths of learned neural heuristics with those of dynamic programming algorithms~\cite{kool2021deep} and the improvement methods~\cite{chen2019learning}~\cite{hottung2019neural}~\cite{da2020learning} and~\cite{lu2019learning}.
Many improvement methods have been developed recently ~\cite{hottung2019neural}~\cite{chen2019learning}~\cite{wu2019learning}~\cite{da2020learning}~\cite{lu2019learning}.  LHI~\cite{wu2019learning} learns which combinations of local operators to apply to solve RPs. RL2OPT~\cite{da2020learning} proposed a deep RL approach to learn an LS heuristic based on 2-opt operators. Chen et al.~\cite{chen2019learning} focuses on improving heuristics and proposes an RL based improvement approach that chooses a region of a graph representation of the problem and then selects and applies established local heuristics. L2I~\cite{lu2019learning} proposed a combination of LS as an improvement operator and ML algorithm. To escape the local minimum, they further improved the solution by a perturbation operator in L2I~\cite{lu2019learning}. 

LS methods can easily get stuck in local minimum~\cite{hansen2006first}. In the OR field, various meta-heuristics have been proposed to avoid local optima. They show promising results to solve RPs~\cite{osman1996meta}~\cite{gendreau2005metaheuristics}. However, meta-heuristics still require expert knowledge.  Our work is similar to L2I~\cite{lu2019learning}, but differs in two important aspects; we use a different set of improving LS operators and we do not use perturbation operators to avoid local minimum. Hence, to produce promising results, we refine LS operators for routing problems and to tackle the local minima problems, we propose to include a set of penalty terms~\cite{voudouris1999guided} that will adaptively guide the LS operators. 
\section{Preliminaries}
In this section we first describe notation used in our approach, and we provide a formal definition of TSP, CVRP, and LS operators used in our approach.
\subsection{Notations}
\begin{table*}[h]
\small
\centering
\caption{Summary of Symbols.}
\begin{tabular}{|l|l|}
\hline
Symbols & Definition\\\hline
n & Nodes \\ \hline
m & Coordinates of nodes\\ \hline
$\pi$& Permutation/Solution  \\\hline
$L(\pi)$ & Objective function \\ \hline
$\pi_L$& Local minimum\\\hline
$h(\pi)$ & Augmented Objective function \\ \hline
$M$  & Maximum roll out steps until \\ \hline
$I$ & Improvement steps  \\ \hline
S  & State \\\hline
A  & Action  \\ \hline
G_i & Demand for customer sites for CVRP \\ \hline
C_i & Capacity of vehicles for CVRP \\ \hline
$p_\theta(A|S)$  &Probability distribution \\ \hline
\end{tabular}
\label{tabile:notation}
\end{table*}

\subsection{TSP}
Let there be $N$ nodes, each representing a city, and a matrix $D=[d_{i,j}]$, which gives the distance between two cities $i$ and $j$. The goal in TSP is to find a sequence of cities (tour) that visits each city exactly once and the total inter-city distance is of minimum length. A tour can be represented as a permutation $\pi$ on the $N$ cities if we interpret $\pi(i)$ to be the next city visited after city $i$ in the tour, $i = 1,\cdot \dotsm \cdot,N$, and $\pi(N)=1$.  The objective of TSP can be written as:
\begin{equation}
\small
\begin{aligned}
L(\pi) = \sum_{i=1}^{N} d_{i,\pi(i)} 
\label{equation:cost}
\end{aligned} 
\end{equation} 
\subsection{CVRP}
In CVRP, there is a depot and a set of N customers. Each customer $i,i \in {1;\cdots;N}$, has a demand $G_i$ to be satisfied. There is a vehicle that makes a number of trips from the depot and serves a number of customers until the vehicle's total demand exceeds the vehicle's capacity $C$, from which the vehicle has to return to the depot. Similar to the TSP problem, the distance between two customers $i$ and $j$ is denoted by $[d_{ij}]$. The traveling cost $[d_{ij}]$ is the cost of a vehicle going from node i to j, with $i; j \in V = {0; 1;\cdots;N}$. Here, the depot is denoted by node 0 for convenience. The goal is to schedule a number of trips that has the minimal total trip distance to serve all customers and respecting the vehicle capacity constraints for each trip and can be mathematically written as:
\[ 
min \sum_{i \in V} \sum_{j \in V} d_{ij}x_{ij}
\label{equation:mrp}
 \]

subject to, \\
$\displaystyle\sum_{i \in V} x_{i,j}  = 1$   $\forall_{j} \in V  \setminus  \{0\}\cdots\cdots\cdots\cdots 1$,\\
$\displaystyle\sum_{j \in V} x_{i,j}  = 1$  $\forall_{i} \in V \setminus  \{0\}\cdots\cdots\cdots\cdots 2$, \\
$\displaystyle\sum_{i \in V} x_{i,0}  = K\cdots\cdots\cdots\cdots 3$, \\
$\displaystyle\sum_{j \in V} x_{0,j}  = K\cdots\cdots\cdots\cdots 4$,\\
             
$u_i - u_i + C x_{ij} \leq C-G_j, \forall_{i,j} \in V \ \{0\}, 
i\neq j \\ $
s.t $G_i + G_j \leq C \cdots\cdots\cdots\cdots 5$    

$G_i \leq u_i \leq C, \forall_i \in V \setminus \{0\} \cdots\cdots\cdots\cdots 6$

$x_{ij} \in {0,1}, \forall_{ij} \in V$

Where K is the number of vehicles available (without loss of generality, it is assumed that K = N for the CVRP we consider), constraints (1) and (2) specify that each customer is visited exactly once, while constraints (3) and (4) specify the in and out degree of the depot, respectively. Constraints (5) and (6) implies the vehicle capacity requirements.

\subsection{Local Search Operators}\label{sec:local}
LS improve feasible solutions through a search procedure, i.e., it starts with an initial feasible solution and replaces a previous solution with a more optimal solution.  Figure~\ref{fig:operators} gives an illustration of LS operators used in our model. The details of operators applied on routing problems are as follows:

\subsubsection{2-opt} Croes et el~\cite{croes1958method} first introduced the 2-optimisation (2-opt) method, which is a simple and commonly used operator. The idea of 2-opt is to exchange the links between two pairs of subsequent nodes. 
Figure~\ref{fig:sfig4} depicts an example of 2-opt.

\subsubsection{Relocate}~\cite{gendreau1992new} is the process where a selected node (target) is moved from its current position in the tour to another position (destination). Hence, the position of the selected node is relocated. Each relocation of a node produces one outcome. Relocate operator is presented in Figure(\ref{fig:sfig5}).

\subsubsection{Swap}~\cite{resende2007fast} Node Swap is another simple optimisation heuristics: Node swap operator exchanges two locations. Node Swap move is a special case of two subsequent 2-opt moves: the first including both cities and the second without them. It involves removing four links and adding four new links. Therefore it is a specific type of 4-opt move. Swap operator is illustrated in Figure~\ref{fig:sfig6}.

\subsubsection{Three permutations}
\begin{figure*}
  \begin{subfigure}[b]{0.45\textwidth}
  \centering
  \includegraphics[width=\linewidth, height = 5cm]{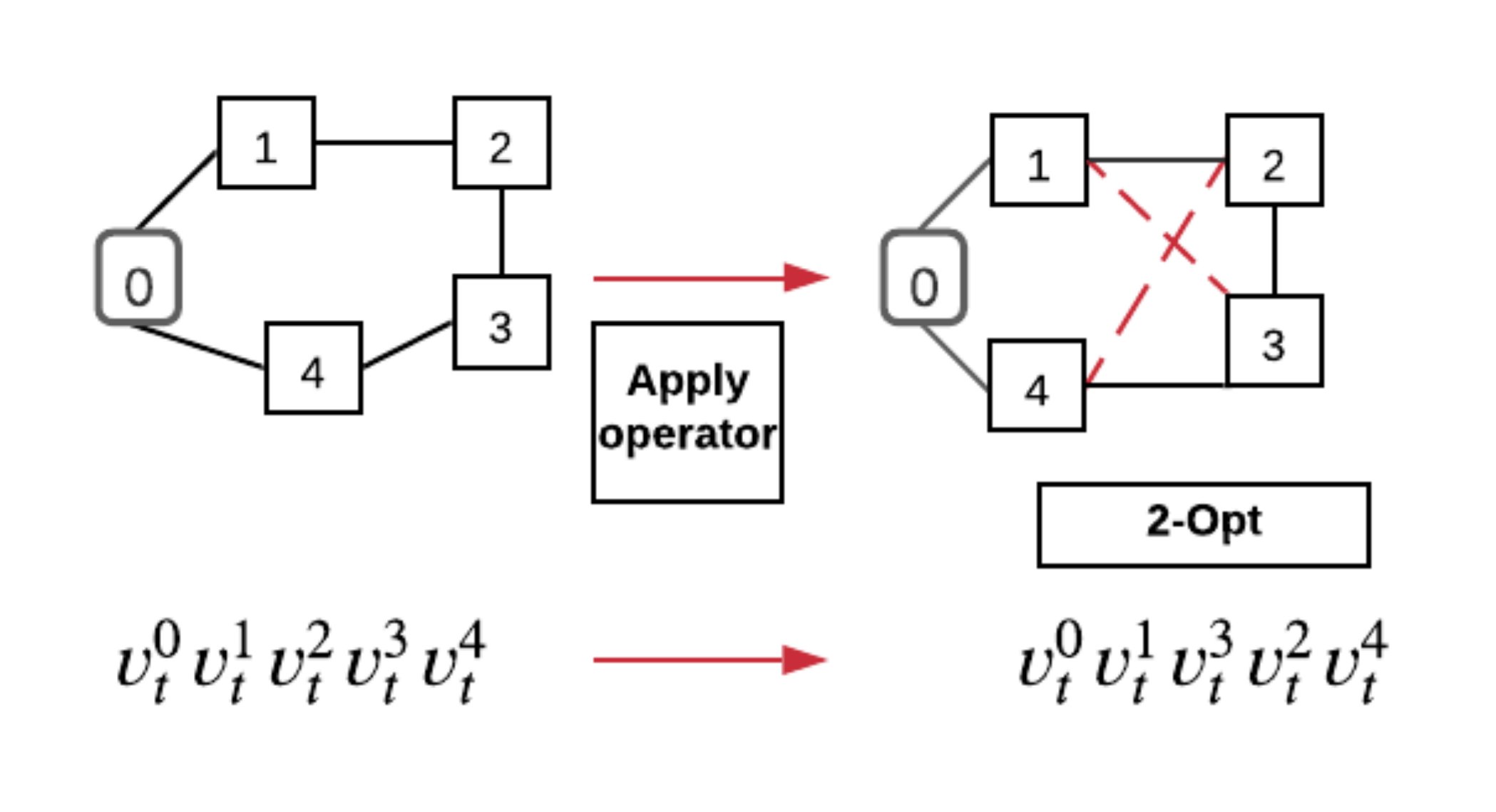}
  \caption{After applying operator 2-Opt, route changes to a new solution with red dashed lines.}\label{fig:sfig4}
  \end{subfigure}
  \hfill
  \begin{subfigure}[b]{0.45\textwidth}
  \centering
  \includegraphics[width=\linewidth, height = 5cm]{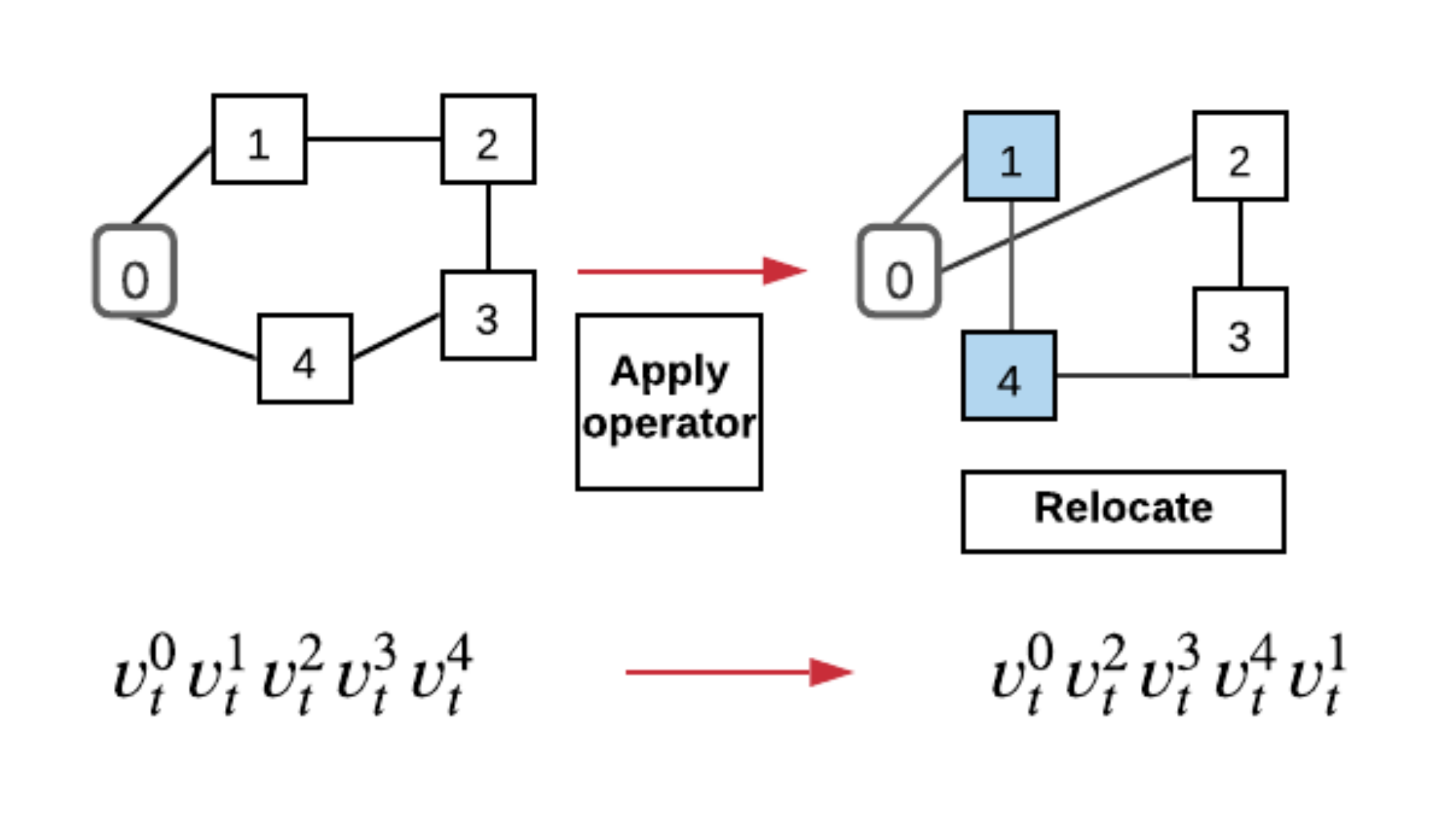}
  \caption{After applying relocate operator, location puts one location after another location $(v_1, v_4)$.}
 \label{fig:sfig5}
  \end{subfigure}
  \centering
  \hfill
  \begin{subfigure}[b]{0.35\textwidth}
  \centering
  \includegraphics[width=\linewidth,height = 5cm]{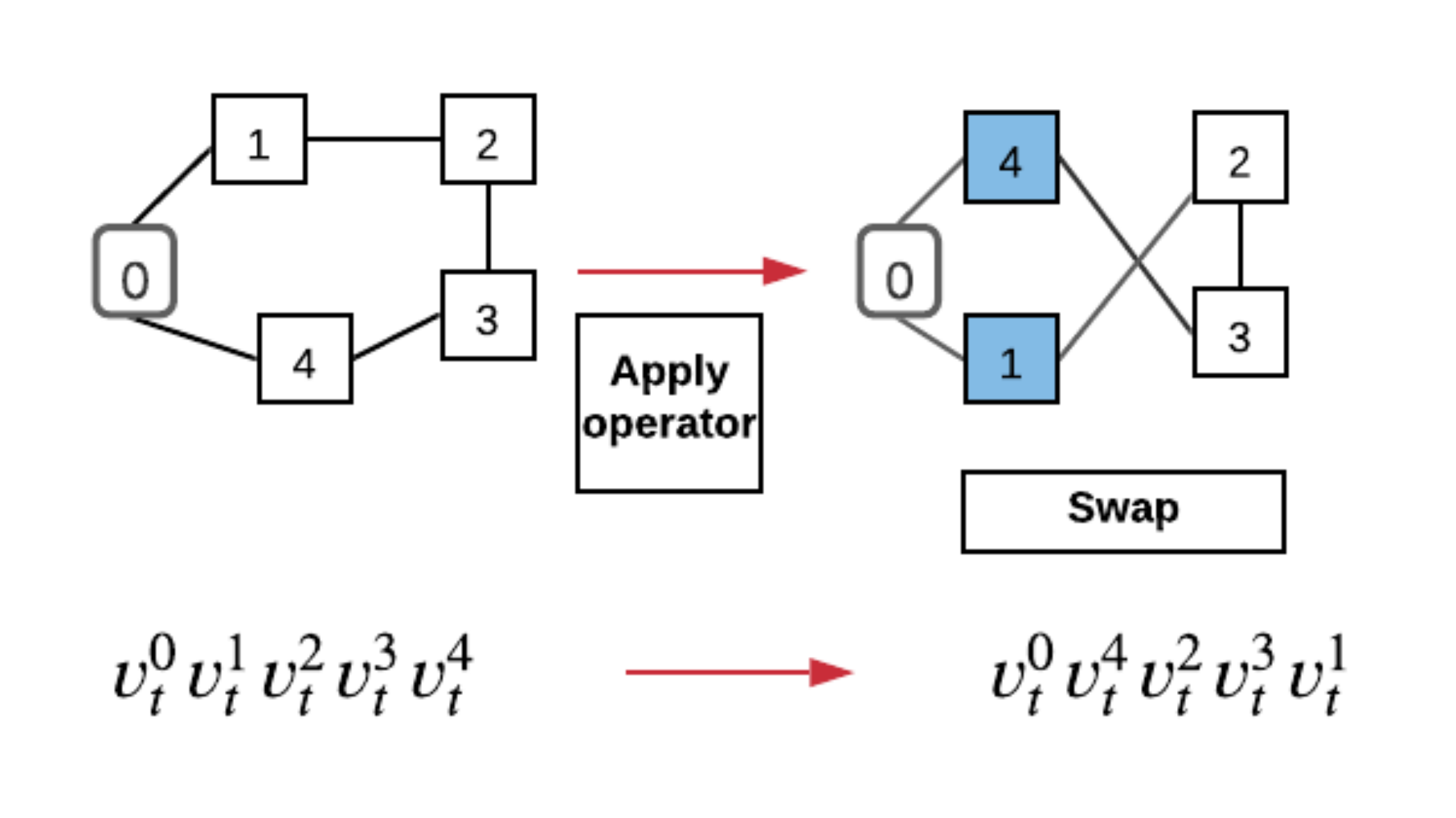}
  \caption{After applying swap operator, location exchanges {$(v_1, v_4)$}}\label{fig:sfig6}
  \end{subfigure}
  \hfill
  \begin{subfigure}[b]{0.55\textwidth}
  \centering
  \includegraphics[width=\linewidth,height = 5cm]{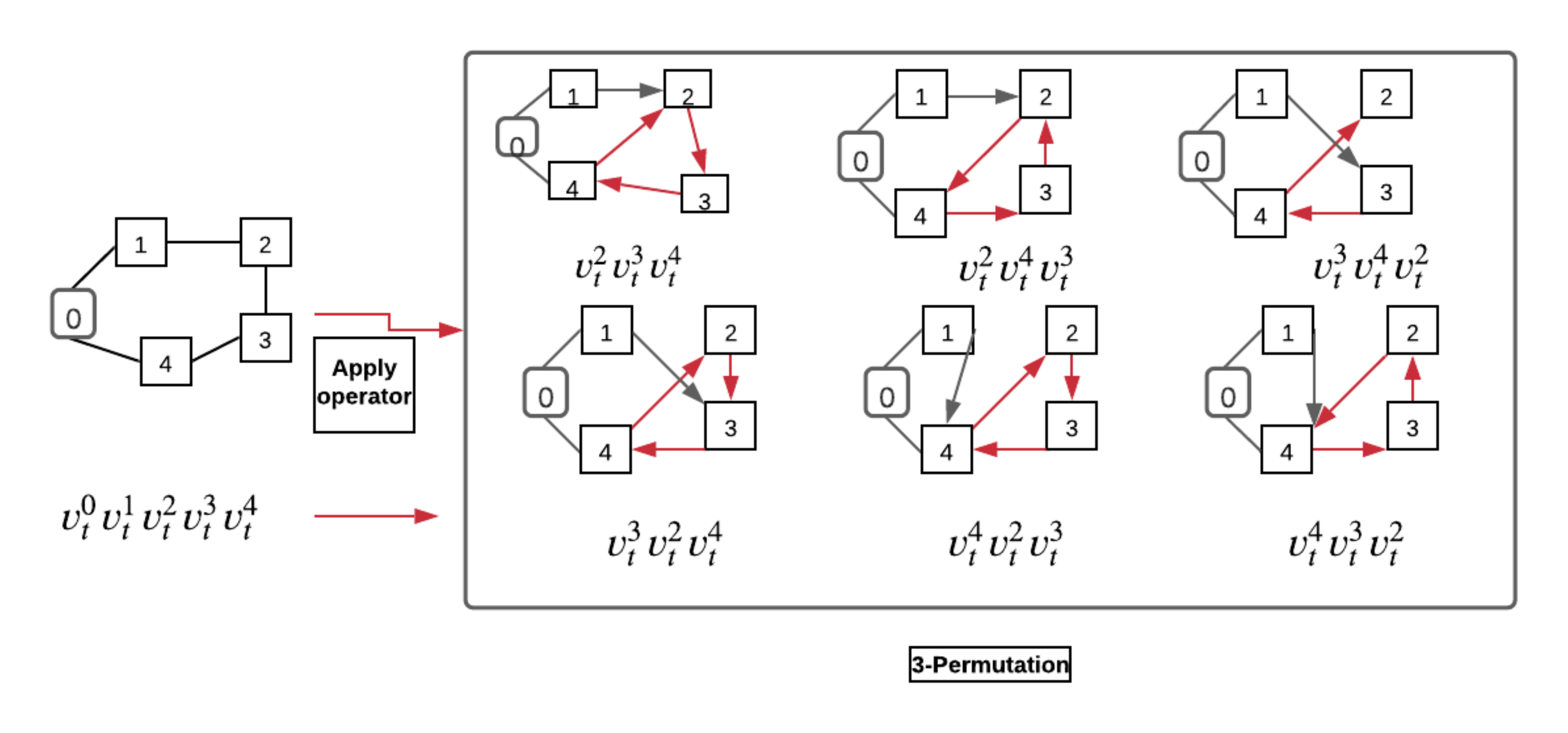}
  \caption{Operator: Three Permutation}
 \label{fig:sfig7}
  \end{subfigure}
 \caption{An illustration of LS operators.}\label{fig:operators}
\end{figure*}

~\cite{sengupta2019local} idea is that three nodes can generate six different orders. Therefore given an existing sequence of three nodes, can create five new solutions. Figure~\ref{fig:sfig7} shows that changing the original order of three nodes can result in a better solution.


\begin{figure*}[!h]
  \vspace{-0.2cm}
  \centering
  \includegraphics[width=0.80\textwidth]{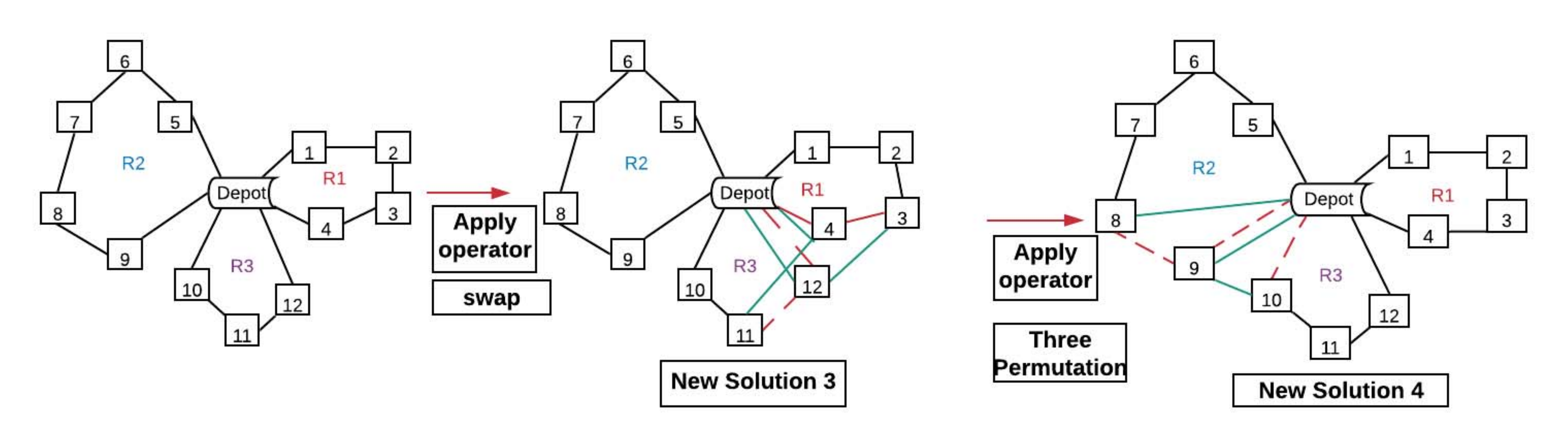}
 \caption{An illustration of a problem instance after applying a set of operators for two routes. Current solution changes to a new solution with dashed lines replaced by red line for the problem CVRP respectively.}
 \label{fig:ope}
  \vspace{-0.1cm}
\end{figure*}

\section{Learning to Guide Local Search (L2GLS)}\label{section:pis}
\begin{figure*}[!h]
  \vspace{-0.2cm}
  \centering
  \includegraphics[width=0.75\textwidth]{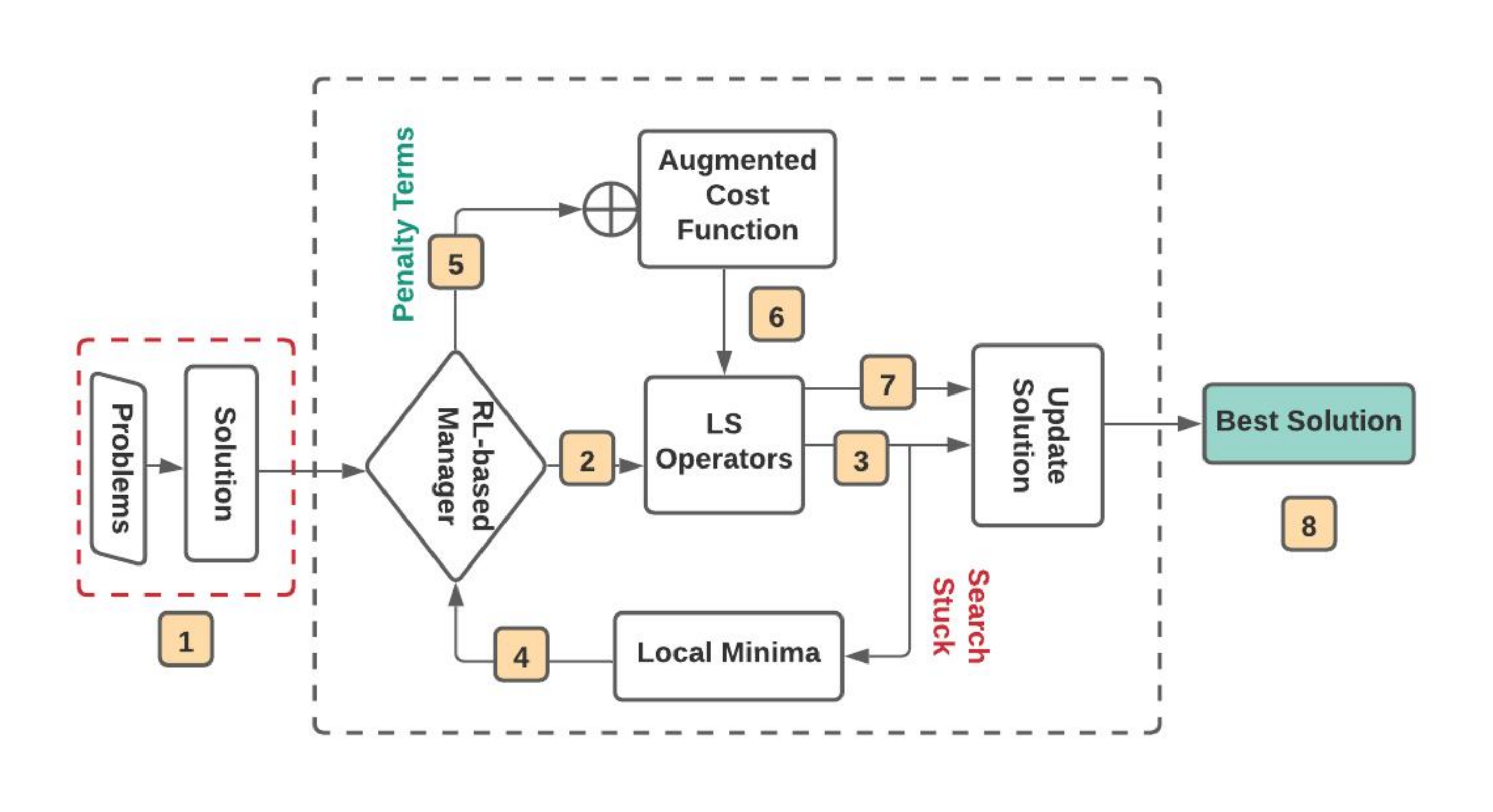}
  \caption{Overview of the Learning to Guide LS Algorithms. L2GLS hierarchy framework. \textbf{1:} Given a problem instance, our algorithm first generates a feasible solution. \textbf{2:} RL-based manager selects a search operator. \textbf{3:} Update the solution
  \textbf{4:} After $I$ improvement steps if no reduce in travelling length means LS stuck in a local minimum. \textbf{5:} The RL-based  manager again take control to take decision. \textbf{6:} RL-based chooses to include a set of penalty terms. \textbf{7}: LS is called again to minimise the cost function. \textbf{8}: After a certain number of steps (M steps), the model choose the best solutions.}\label{fig:pisalgo}
  \vspace{-0.1cm}
\end{figure*}

This section introduces our “Learning to Guide Local Search” (L2GLS) algorithm. Figure~\ref{fig:pisalgo} illustrates an overview of our approach. Our algorithm starts with a feasible solution and continuously improves the solution using LS operators. L2GLS exploits problem and search-related information to escape local minimas by augmenting the objective function of the problem with a set of penalty terms. The penalty terms are adaptively adjusted and reflects the degree that a solution feature causes a solution to be sub-optimal and candidate for further local improvement. This avoids ad-hoc perturbations to escape local minimal. We found this strategy is able to converge to good solutions faster.  

Our method follows threshold-based rules to decide whether we should continuously improve the current solution using LS operators or apply the penalty term. If our RL-based manager decides that the current solution could still be improved, it chooses one LS operator to improve it. When no objective value reduction is made for a pre-determined 
$I$ number of applied actions (which are operators applied), we consider that the LS has reached a local optimal and the RL-based manager chooses to use penalty term to escape this local optimal. If after M times of reaching a local optimal and applying the penalty terms still result in no cost reduction across these M runs, then the algorithm and search terminates and the best solution of these runs is returned. Having described our overall method, we are now ready to present the details of our RL-based manager. 

\subsection{Reinforcement Learning based Manager}
The main component of L2GLS is a RL based manager that guides the search by adaptively selecting which LS operators to use in each iteration and their parameters (e.g, which nodes to apply the operator). It will also detect if there has not been improvement in the solution for a few iterations to determine when we reach a local optima. Then uses the penalty term with adaptive weights to escape the local optima. We will detail these components in the following.

The RL-based manager is a deterministic model, which is defined by the tuple ($S$, $A$, $T$, $R$, $P$), where $S$ is the states, $A$ is the actions, $T$ is the deterministic transaction function ($T: S \times A \rightarrow S)$, $R$ is the reward function ($R: S \times A \rightarrow \mathcal{R})$ and $\pi$ is the policy ($\pi: S \rightarrow A$). We define each of this and relate them to the optimisation problem.

\subsubsection{States}\label{subsubsection:state}
Each state represents the problem instance and a solution, and made up of problem-specific and solution-specific features. Problem-specific features include the node location, and demand of each node for CVRP, and they are considered problem-specific because they are associated with the problem and invariant across solutions. Solution-specific features are based on the given solution. For each node, we compute its neighbouring nodes that are visited before and afterwards along with their relevant distances. State also includes recently taken actions and the effects of the actions. We follow the L2I~\cite{lu2019learning} approach for the state. Next, we describe the  features. First, we introduce the description of all the features for TSP, where the location of node $i$ is represented by ($m_i, n_i$), for a 2D location, but this can be generalised to more than 2 dimensional spaces. Previously visited location of node $i$ is denoted by ($m_{i^-}, n_{i^-}$) and the location visited after $i$ is ($m_{i^+}, n_{i^+}$). The neighbouring nodes distance from $i^-$ to $i$ is $(d_{i^-,i})$, distance from $i$ to $i^+$ is $(d_{i,i^+})$, and distance from $i^-$ to $i^+$ is $(d_{i^-,i^+})$. The state also includes the action taken previously and the effects caused by taking the step, i.e., the actions taken $h$ steps before, denoted as $a_{t-h}$, and their effects is denoted by $e_{t-h}$. For example, $a_{t-h}$, $1\leq h \leq H$, is the action taken h steps before current step t, and its effect $e_{t-h}$ is +1 if the action led to a reduction of total distance, -1 otherwise. All the features for CVRP are the same as TSP, except along with the customer location ($m_i, n_i$), we have additional tuple dimensions of the demand of each customer $i$, denoted by $G_i$, and the free capacity of the route containing customer $i$ denoted as $C_i$.
\subsubsection{Actions}\label{section:action}
L2GLS actions involve selection of appropriate LS operators and their operands to apply to current solution to improve it. Each action is a tuple $(o, \phi)$, where $o \in O$ are the set of LS operators and $\phi \in \Phi$ is the parameters associated with the operators, e.g., remove two edges and reconnect their endpoints in the 2-opt operator (note that $\phi$ will be different for each operator). For a given problem, the same operator with different parameters may perform differently. Consequently, each operator with different parameters act as separate actions and we let the RL model learn how to use them best. Please refer to the Supplementary material to see a description of how the LS operators work.

\subsubsection{Transition Function and Policy Networks}\label{section:tf}
The transition function is deterministic, i.e., given current state and an action, it will always transit to only one next state. However in terms of learning a policy to guide the actions, the state and action spaces are very large, and it is not feasible to be able to see every possible combination of these to train the policy. Hence we adopt a policy gradient approach to do so, i.e., use a neural network to learn and represent the policy. In a policy gradient approach, a neural network, also known as a policy network, represents the policy, i.e., the mapping function from input states to output actions. 
This network is defined by a set of hyper-parameters $\theta$ and outputs a probability distribution of actions given an input state. Given a state $S$, our model defines a probability distribution $P_\theta(A|S)$, from which we can sample actions to obtain a solution tour $\pi$. 
In order to train our policy network, we define the loss of the policy as $\mathcal{L}(\theta|S)=E_{\pi \sim P_{\theta}(.|S)}[L(\pi|S)]$ $\forall a \in A$ (where $.$ is over all actions), which is the expectation of solution $(\pi)$ given the probability distribution of $p_\theta(.|S)$ from which we can obtain actions given state. We optimise $\mathcal{L}$ by gradient descent, using the REINFORCE~\cite{williams1992simple} gradient estimator with baseline $b(S)$:
\begin{equation}
   \nabla \mathcal{L}(\theta|S) = E_{\pi \sim P_{\theta}(.|S)}[ (L(\pi|S) - b(S)))\nabla_{\theta} log P_{\theta}(\pi|S)]
\label{equation:ameq}
\end{equation}
We sample solutions and compare it with baseline $b(S)$. We use baseline $b(S)$ to reduce gradient variance (adjusting the probability proportion to result we get compare to the baseline) and speed the learning process. We use ADAM~\cite{kingma2014adam} as optimiser. Problem-specific and solution-specific states are transformed into an embedding. The embeddings are then fed to the attention network (see Supplementary for the overview of the policy network). In the policy network, we first fed the problem-specific states to the attention network as an embedding, then the output of the attention network is concatenated with a sequence of recent actions and their effects (solution specific states). Lastly, the output of the attention network is fed into a network of two fully connected layers, producing action probabilities. 
\subsubsection{Rewards}\label{section:reward}
We experiment with two reward functions that were also used in~\cite{lu2019learning}. The first reward is if an selected action (operator and parameters) results in a reduction of the objective value of the current solution, the reward is +1; otherwise, -1. The second reward is defined over a larger range of values. During the first improvement iteration, the objective value attained for the solution is taken as a baseline. For each subsequent iteration, the reward for an operator is equal to the difference between the objective values achieved after applying the operator and the baseline. Similar to~\cite{lu2019learning}, we found there is generally a diminishing improvement in latter iterations hence we give the same reward for all operators applied in one iteration.  
\subsection{Penalty Term}\label{section:pt}
The penalty term aims to identify solution features that contribute significantly towards the overall cost of a solution and encourage their replacement in the tour. It does this by adding penalty terms to these features.

The penalty term uses two types of information: one is the cost of each feature to the solution and the set of features currently in the solution. This information is transformed into constraints on features which then are incorporated in the cost function using the adjusted penalty terms. Constraints on features are made possible by augmenting the cost function~$L(\pi)$ of the problem to include a set of penalty terms. 

The high-cost features are being penalised. For example, in the TSP, the features are the travel costs between pairs of cities, and generally high intercity distances in a tour are not desirable, though we can not exclude them from the beginning because they may be needed to connect remote cities in the tour. 

When the search is stuck in a local minimum, our RL manager modifies the parameters on the penalty term~\cite{voudouris1999guided} to encourage/discourage certain solution features and subsequently calls on LS to improving the solution using the augmented objective function, defined as follows:
\begin{equation}
\small
\begin{aligned}
 h(\pi) = L(\pi) + \lambda \cdot \sum_{i=1}^{F} p_{i} \cdot I_{i}(\pi)
\label{equation:augmented}
\end{aligned} 
\end{equation}
where $F$ is the number of features for the problem instance, $p_i$ is the penalty parameter corresponding to feature $f_i$ and $\lambda$ is a regularisation parameter for the penalty terms. An indicator function $I_i(\pi)$ indicates whether the solution feature $i$ is part of the solution $\pi$ (hence we only penalise and consider features that are part of the current solution). 
\begin{equation}
\small
\begin{aligned}
I_{i}(\pi) =  \begin{cases}
      1, & \text{solution  $\pi$ has feature $i$}\\
      0, & \text{otherwise}   
    \end{cases} 
\label{equation:au}
\end{aligned} 
\end{equation}

The penalty parameter $p_i$  gives the degree up to which the solution feature $f_i$ is constrained. The regularisation parameter $\lambda$ represents the relative importance of penalties with respect to the solution cost and has great significance because it controls the influence of the information on the search process. L2GLS iteratively uses LS to improve the solution and increments some of the parameter parameters $p_1,\cdot \cdot \cdot,p_F$, each time LS converges to a local minimum. To determine which penalty parameter to increment, the following utility function is computed for each feature:
\begin{equation}
\small
\begin{aligned}
U(\pi_L, f_i)= I_i(\pi_L)\cdot\frac{d_i}{1+p_i}
\label{equation:utility}
\end{aligned} 
\end{equation} 
where $d_i$ is the cost (e.g., intercity travel distance for TSP) of the feature $f_i$. When at a local minimum (denoted as $\pi_L$) recall we seek to penalise those solution features that contribute most towards the current solution. The feature(s) that have the maximal utility have their penalty parameters incremented to penalise them (see Equation \ref{equation:augmented}), which is used by our RL based manager to guide LS operators. But we also do not wish to continually penalise the same features (ones with the largest $d_i$). Hence the utility function is proportional to feature cost $d_i$, but weighted by the inverse of $(1+p_i)$, which gets larger as we penalise a feature more, and subsequently reduces the utility function as that feature is penalised more. This allows other features to be considered over time to be penalised and increases the searching.

Initially, all the penalty parameters are set to 0 (i.e., no features are constrained), and when LS is searching, a call is made to LS to find a local minimum of the augmented cost function using Equation~\ref{equation:augmented}. After each local minimum, the algorithm takes a modified action (with LS operators) on the augmented cost function and uses LS starting from the previously found local minimum. Information is inserted in the augmented cost function by selecting which penalty parameters to increment (e.g., sources of information are the cost of features (edges between two nodes cost) and the local minimum itself). That is how our algorithm offers guidance to the LS.
\section{Evaluation Setup}\label{appendix:ex}
For the two problems we are focusing on in this paper, TSP and CVRP, we follow the evaluation setups of AM~\cite{kool2018attention} and L2I~\cite{lu2019learning}, which are two state-of-the-art approaches for these problems. All reported metrics, such as the final travelling cost and the running time, are always computed as the average over 1000 randomly generated instances/samples. 

\subsection{Problem Specific Setup} For both problems, we generate the training instances where the coordinates of each node are sampled in a uniformly random way from the unit square [0; 1] $\times$ [0; 1]. For TSP, we generated instances with N = 20; 50; 100 nodes and consider these as a small sized TSP problem. Most of the previous work evaluated their models with these sizes, so we seek to first show the performance of L2GLS on small problems. For a large problem, we generated instances with N = 200; 500; 1000 nodes. 

For CVRP, we solve the problems with a setup as described in L2I~\cite{lu2019learning}. The location $(m_i; n_i)$ of each customer, as well as of the depot, is uniformly sampled from a unit square (specifically, $m_i; n_i$ are uniformly distributed in the interval [0;1], respectively), and the travelling cost between two locations $d_{i;j}$ is simply the corresponding distance. The demand $G_i$ of each customer is uniformly sampled from the discrete set $\{ 1,\cdots,9\}$. We consider instances with $N$ = 20; 50; 100 nodes as a small problem, where the capacity of a vehicle is 20, 30, 40 for $N$ = 20; 50; 100, respectively. Similar to TSP, a large problems consists of $N$ = 200; 500; 1000 nodes. We keep the capacity of the vehicle 50 for all sizes of the large problems. 

\subsection{Policy Network Settings}
Our method is directly comparable to L2I~\cite{lu2019learning}, hence to train the model we use the same parameters setting for fair comparison. We use ADAM with a learning rate of 0.001. Unless otherwise stated, we randomly initiate a feasible solution for a problem instance and a given policy and then iteratively update the solution following the policy $M$ = 40000 times. After $I$ = 6 consecutive improvement steps, we use the penalty term. In After a maximum number of rollout steps, the algorithm stops and among all the 40000 visited solutions, we choose the best as the final solution for a given problem instance. We selected the values of $\lambda$ as constant (in the equation~\ref{equation:augmented}), because with the constant value 0.3 was recorded best performance during training. Performance metrics used in the paper includes total travelling cost/distance and the computational running time, which are computed as the average over 1000 random instances. For encouraging exploration, we use a greedy~\cite{sutton2018reinforcement} approach, where the RL manager will choose a random improvement action with a probability of 0.05 following~\cite{lu2019learning}. Lastly, L2GLS was implemented in Python, on a workstation with an Intel Xeon 2.4 GHz CPU with 56 cores.

 \begin{table}[ht]
\scriptsize
\centering
\caption{Experiment results on small TSP. Average tour length, gap percentage (lower is better, bold is best). Results with * are reported from other papers. \textbf{($**$)} means the value reported for the problem size outperformed all the methods, including the optimal one for TSP. \textbf{($-$)} means Bresson et al.~\cite{bresson2021transformer} method never tested on TSP20 instance.}
\begin{tabular}{|l|ll|ll|ll|}
\hline
Method & \multicolumn{2}{|c|}{TSP20} & \multicolumn{2}{|c|}{TSP50}& \multicolumn{2}{|c|}{TSP100} \\ \hline
\textbf{Solver}    & TourL  & Gap & TourL  &  Gap & TourL   & Gap\\\hline
Concorde~\cite{applegate2006concorde}      & 3.84  & 0.00\%  & 5.70  & 0.00\% & 7.77  & 0.00\%\\ \hline
\textbf{Heuristics} &  &  & &  &  &     \\\hline
LKH3~\cite{helsgaun2017extension} & 3.84  & 0.00\%  & 5.70 & 0.00\% & 7.77 &0.00\%\\ \hline
\textbf{Constructive (SL)} &  &  & &  &  &     \\\hline
GCN*~\cite{joshi2019efficient}& 3.86  & 0.52\% & 5.87   & 2.98\% & 8.41 & 8.23\%\\
NETSP-Net \cite{sultana2020learning} & {3.85}  & 0.26\%  & {5.85}  & 2.63\% & {8.31}  & 6.94\% \\
Fu et al.*~\cite{fu2020generalize} & 3.83  & ** & 5.69 & ** &{7.76} & **\%\\\hline
\textbf{Constructive RL)} &  &  & &  &  &    \\\hline
AM~\cite{kool2018attention}& 3.87 &0.78\%  & {5.80}&1.75\%   & {8.15}  &4.89\%\\
ERRL~\cite{sultana2020learningb}& 3.83 &0.78\%  & {5.73}&1.75\%   &{7.85}  &4.89\%\\
POMO\cite{kwon2020pomo} & 3.83  & ** & 5.73 & 1.75\% &{7.84} & 7.07\%\\
Bresson et al.*~\cite{bresson2021transformer} & -  & - & 5.69 & ** &{7.81} & 7.07\%\\
\textbf{Improvement} &  &  & &  &  &    \\\hline
LHI*\cite{wu2019learning} & 3.83 & **  & 5.74   & 0.71\% & 8.01 & 3.08\%\\
L2I. \cite{lu2019learning} & 3.84  & 0.00 & 5.72   & 0.7\% & 7.90 & 1.67\%\\
RL2OPT.*\cite{da2020learning} & 3.84  & 0.00 & 5.72   & 0.7\% & 7.91 & 1.80\%\\\hline
\textbf{L2GLS(Ours)} & \textbf{3.72} & ** & \textbf{5.65} & ** & \textbf{7.69} &**\\\hline
\end{tabular}
\label{tabile:tsp}
\end{table}

\section{Experiments and Analysis}\label{section:5}
The evaluation procedure used in this study involves:
 \begin{itemize}
 \item  We evaluated our method using the large instance to show the method's scalability. Few papers considered the large scale instances for TSP and CVRP;
 \item The model is evaluated using various small scale random instances, as previous neural network-based approaches typically focus on randomly generated small scale data.
 \item Computational running time analysis for small and large scale instances.
 \item To demonstrates generalisation ability, we compare the quality of solutions of our method with the benchmark TSP instances from the TSPLIB library~\cite{reinelt1991tsplib}, where we tested `thirty-nine' Euclidean instances. For the first time, in the L2O field, we evaluated our model up to 1002 benchmark instances, achieving close to the optimal solution. Moreover, we also compare the solution quality of our method using various scenarios of CVRP benchmark instances from~\cite{uchoa2017new}. Demonstrate the model generalisation using various training and testing sizes of instances for TSP and CVRP. Due to space limitations, we presented these results in a supplementary document.
\item We illustrate the impact of L2GLS variants and evaluated on randomly generated instances for TSP and CVRP. Moreover, we compare L2GLS performance with and without using penalty terms on TSPLIB~\cite{reinelt1991tsplib}.
\end{itemize}

\begin{table}[ht]
\scriptsize
\centering
\caption{Experiment results on small CVRP. Average tour length and gap percentage (lower is better, best in bold). Results with * are reported from AM~(\cite{kool2018attention}) and Chen et al.~\cite{chen2019learning}. \textbf{($**$)} means the value reported for the problem size outperformed all the methods, including the optimal one for CVRP.}
\scriptsize
\begin{tabular}{|l|ll|ll|ll|}\hline
Method & \multicolumn{2}{|c|}{CVRP20} & \multicolumn{2}{|c|}{CVRP50}& \multicolumn{2}{|c|}{CVRP100} \\ \hline
\textbf{Solver}    & TourL  & Gap(\%)& TourL & Gap(\%)  & TourL & Gap(\%)  \\\hline 
LKH3~\cite{helsgaun2017extension} & 6.14  &  0.00   & 10.39 & 0.00\%  &  15.67 &0.00  \\ \hline
\textbf{Heuristics}& &  &   &  &  &  \\\hline
Or-tools* &  6.43& 4.73   &11.43 & 10.00   & 17.16& 9.50   \\\hline
\textbf{Constructive} &  &  & &  &  &  \\\hline
AM~\cite{kool2018attention}& 6.67 & 8.63 & 11.00  & 5.87  & {16.99} &8.42\\
Nazari et al.~\cite{nazari2018reinforcement}& 7.07& 15.14  & 11.95& 15.01 &  17.89 &14.16  \\
POMO\cite{kwon2020pomo}& 6.35 &3.42  &10.74& 3.36    & 16.15 &3.06 \\
ERRL~\cite{sultana2020learningb} & {6.34} & 3.25  & {10.77} & 3.65 & {16.35} & 4.02 \\\hline
\textbf{Improvement} &  &  & &  &  &   \\\hline
Chen et al.~\cite{chen2019learning} & 6.16& 0.3 & 10.51& 1.15 &  16.10  &2.72 \\
LIH.~\cite{wu2019learning} & 6.16  & 0.26 & 10.72  & 0.35 & 16.30 & 1.60\\
L2I~\cite{lu2019learning} & 6.12 & **  & {10.35} & **  & {15.57} & ** \\\hline
\textbf{L2GLS (Ours)} &\textbf{5.85}  & **  & \textbf{10.30} & **  & \textbf{14.67} &**\\\hline
\end{tabular}
\label{tabile:cvrp}
\end{table}
\begin{table*}[h]
    \caption{Experiment results of  average tour length on large TSP and CVRP (lower is better, best in bold).}
    \scriptsize
     \centering
     \begin{subtable}{0.45\linewidth}
    \begin{minipage}{0.45\linewidth}
      \caption{TSP}
\begin{tabular}{|l|l|l|l|}\hline
Method & \multicolumn{1}{|c|}{TSP200} & \multicolumn{1}{|c|}{TSP500}& \multicolumn{1}{|c|}{TSP1000} \\ \hline
\textbf{Solver}    & TourL   & TourL  & TourL   \\\hline 
Concorde~\cite{applegate2006concorde}~ & 10.15 & 16.542  & 23.130  \\\hline
L2I~\cite{lu2019learning} & 10.95  & 17.68  & 26.13  \\
\textbf{L2GLS} & \textbf{10.10}  & \textbf{16.12} & \textbf{22.37} \\\hline
        \end{tabular}
        \label{tabile:TVlargetsp}
    \end{minipage}%
    \end{subtable}%
     \begin{subtable}{.45\linewidth}
    \begin{minipage}{.45\linewidth}
      \centering
        \caption{CVRP}
        \begin{tabular}{|l|l|l|l|}
            \hline
Method & \multicolumn{1}{|c|}{CVRP200} & \multicolumn{1}{|c|}{CVRP500}& \multicolumn{1}{|c|}{CVRP1000} \\ \hline
\textbf{Solver}& TourL  & TourL & TourL \\\hline 
L2I~\cite{lu2019learning} &  {28.4}  & {64.68} & {128.49} \\
\textbf{L2GLS} &\textbf{24.69}   & \textbf{60.67}  & \textbf{124.29} \\\hline
        \end{tabular}
        \label{tabile:TVlargetsp1}
    \end{minipage} 
    \end{subtable}%
\end{table*}
\subsection{Performance analysis: Small scale TSP and CVRP instances}~\label{section:result} 
In this section, we report the performance of our method on various sizes of small scale instances for TSP and CVRP. 
\subsubsection{TSP}
Table~\ref{tabile:tsp} reports the performance of L2GLS and baseline methods. We have five different baseline approaches in Table~\ref{tabile:tsp}: solver Concorde~\cite{applegate2006concorde}, heuristics~\cite{helsgaun2017extension}, learning methods using constructive heuristics, supervised and RL approaches; and improvement heuristics namely LHI~\cite{wu2019learning}, RL2OPT~\cite{da2020learning} and L2I~\cite{lu2019learning}. There are several ML models in the literature, but they all produce poor performance compare to our approach, thus being omitted here. Our approach outperformed all recent learning method, presented in Table~\ref{tabile:tsp}. Note that the L2GLS method produces state-of-the-art results for TSP and to the best of our knowledge is the first learning-based framework that outperforms Concorde~\cite{applegate2006traveling} for the random dataset. 
\subsubsection{CVRP}~\label{section:smallcvrp} 
Table~\ref{tabile:cvrp} reported the performance of L2GLS methods for CVRP, there we compare our method against the current RL algorithms~\cite{nazari2018reinforcement}~\cite{kool2018attention}~\cite{lu2019learning}~\cite{wu2019learning}~\cite{sultana2020learningb}. In particular, the average tour length achieved by L2GLS is substantially shorter than LKH3~\cite{helsgaun2017extension}. L2GLS also outperforms all the neural network-based approaches, including recent works by Chen et al.~\cite{chen2019learning} and L2I~\cite{lu2019learning}.
Our algorithm generates state-of-the-art results for TSP and CVRP for small problems. 
\subsection{Scalability Analysis: Large Scale TSP and CVRP instances}\label{appendix:large}
In this section, we compared the L2GLS performance to a most recent state of the art learning-based approach L2I~\cite{lu2019learning} for large sized problems, as we discussed previously, most state of the art works have not evaluated on due to their lack of scalability. Tables~\ref{tabile:TVlargetsp} and~\ref{tabile:TVlargetsp1} reports the results of large TSP and CVRP instances respectively. 
Our method significantly outperforms L2I~\cite{lu2019learning} for all the TSP and CVRP instances, even for a large number of nodes N=1000. For CVRP1000, we averaged over 200 instances instead of 1000 instances due to time constraint. 

\begin{figure}
\centering
\begin{subfigure}[b]{0.35\textwidth}
\includegraphics[width=\linewidth, height = 5cm]{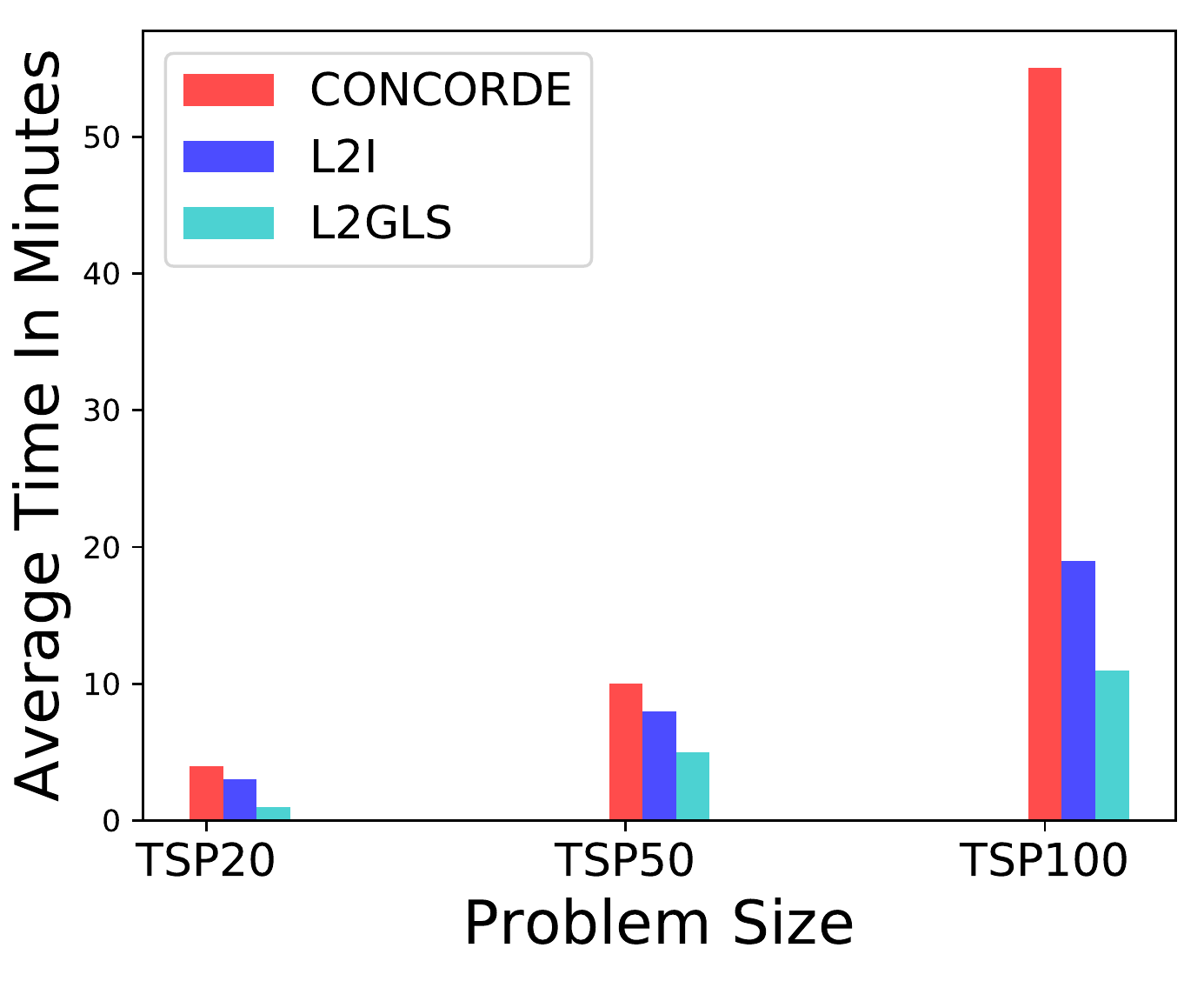}
    \caption{TSP computation time.}\label{fig:tspstime}
  \end{subfigure}
  \hfill 
\begin{subfigure}[b]{0.35\textwidth}
\includegraphics[width=\linewidth, height = 5cm]{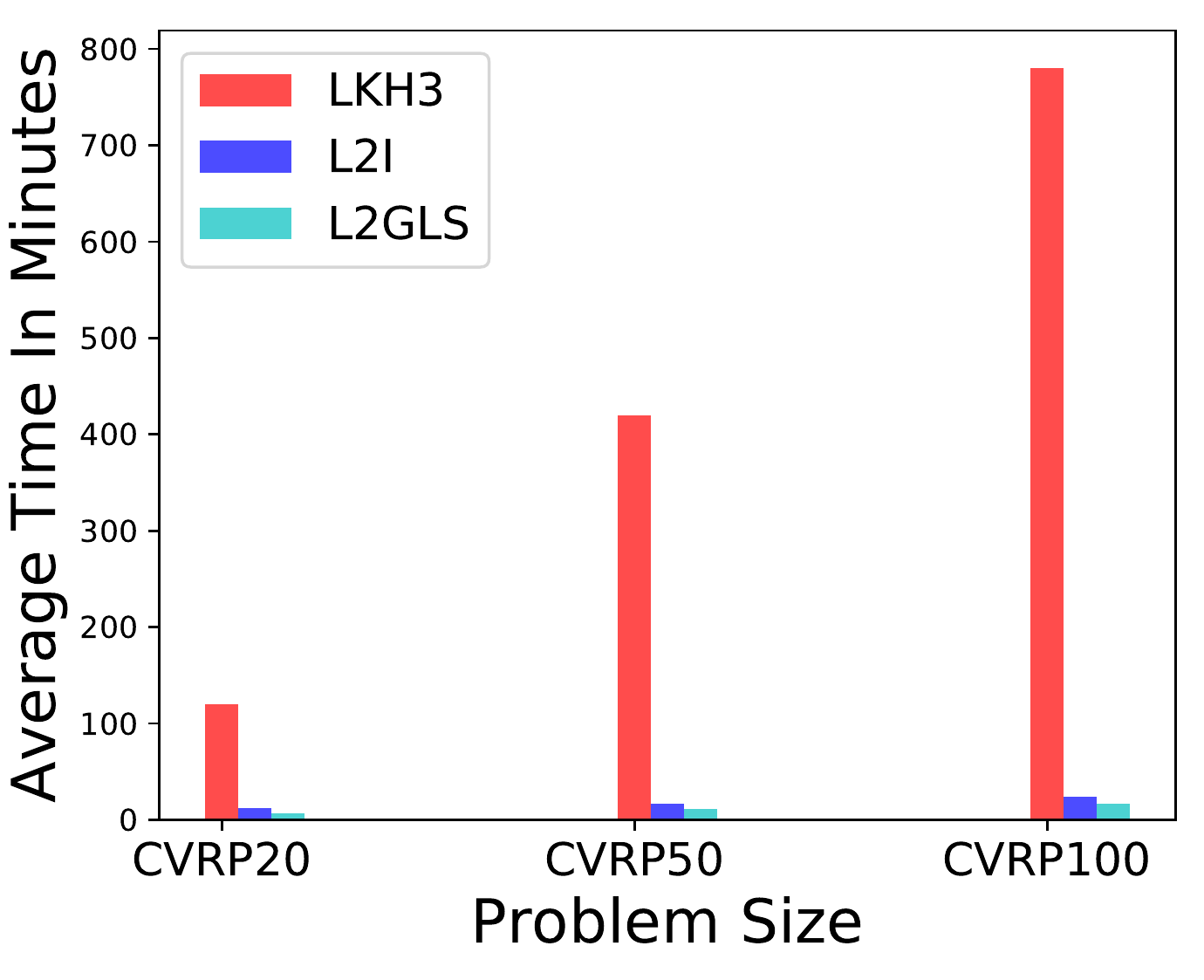}
    \caption{CVRP computation time.}\label{fig:vrpstime}
  \end{subfigure}
   \caption{Average computation time for smaller TSP and CVRP.}\label{fig:timetsp}
   \end{figure}
\begin{figure}
\centering
\begin{subfigure}[b]{0.35\textwidth}
\includegraphics[width=\linewidth, height = 5cm]{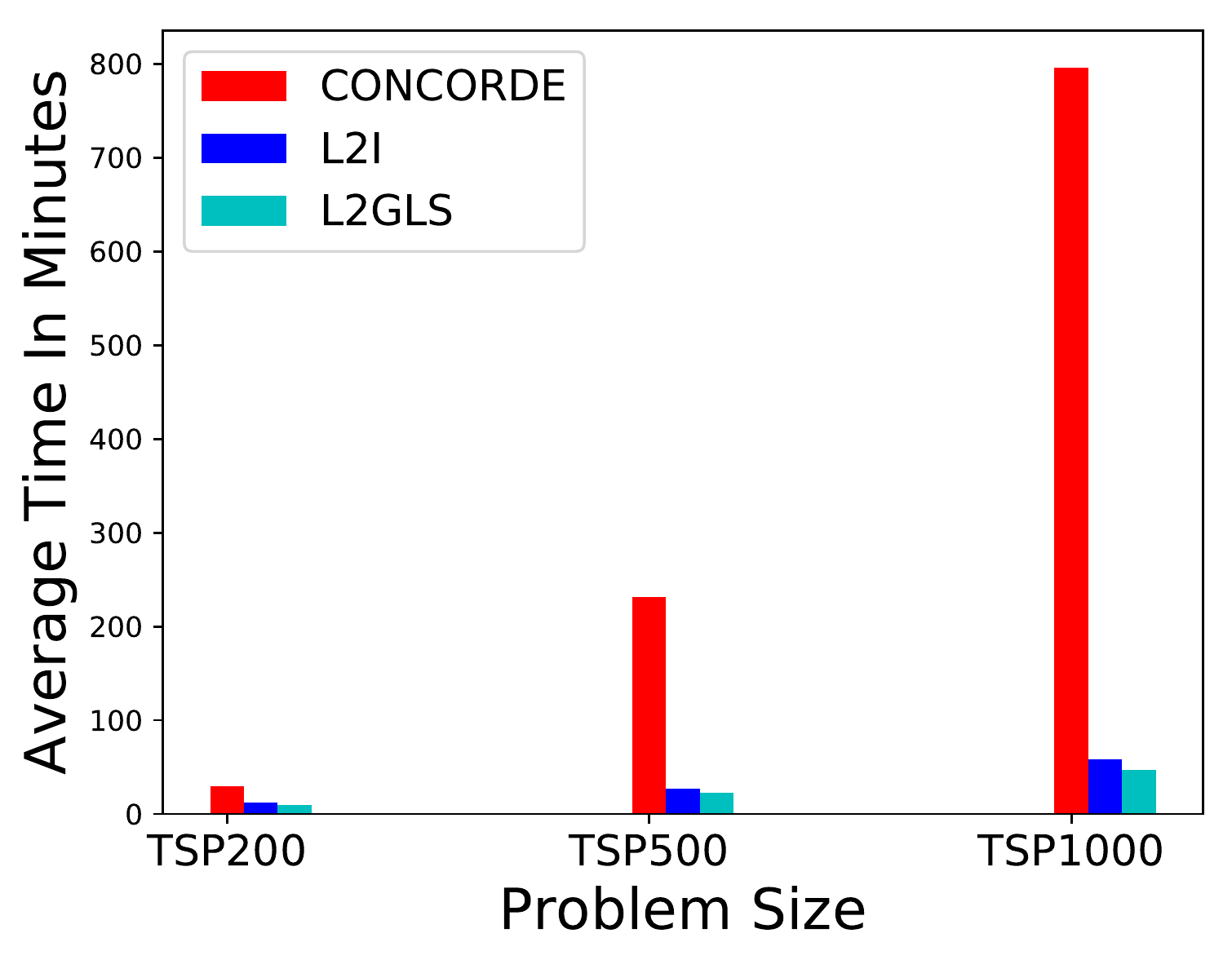}
    \caption{TSP computation time.}\label{fig:tspltime}
  \end{subfigure}
  \hfill 
\begin{subfigure}[b]{0.35\textwidth}
\includegraphics[width=\linewidth, height = 5cm]{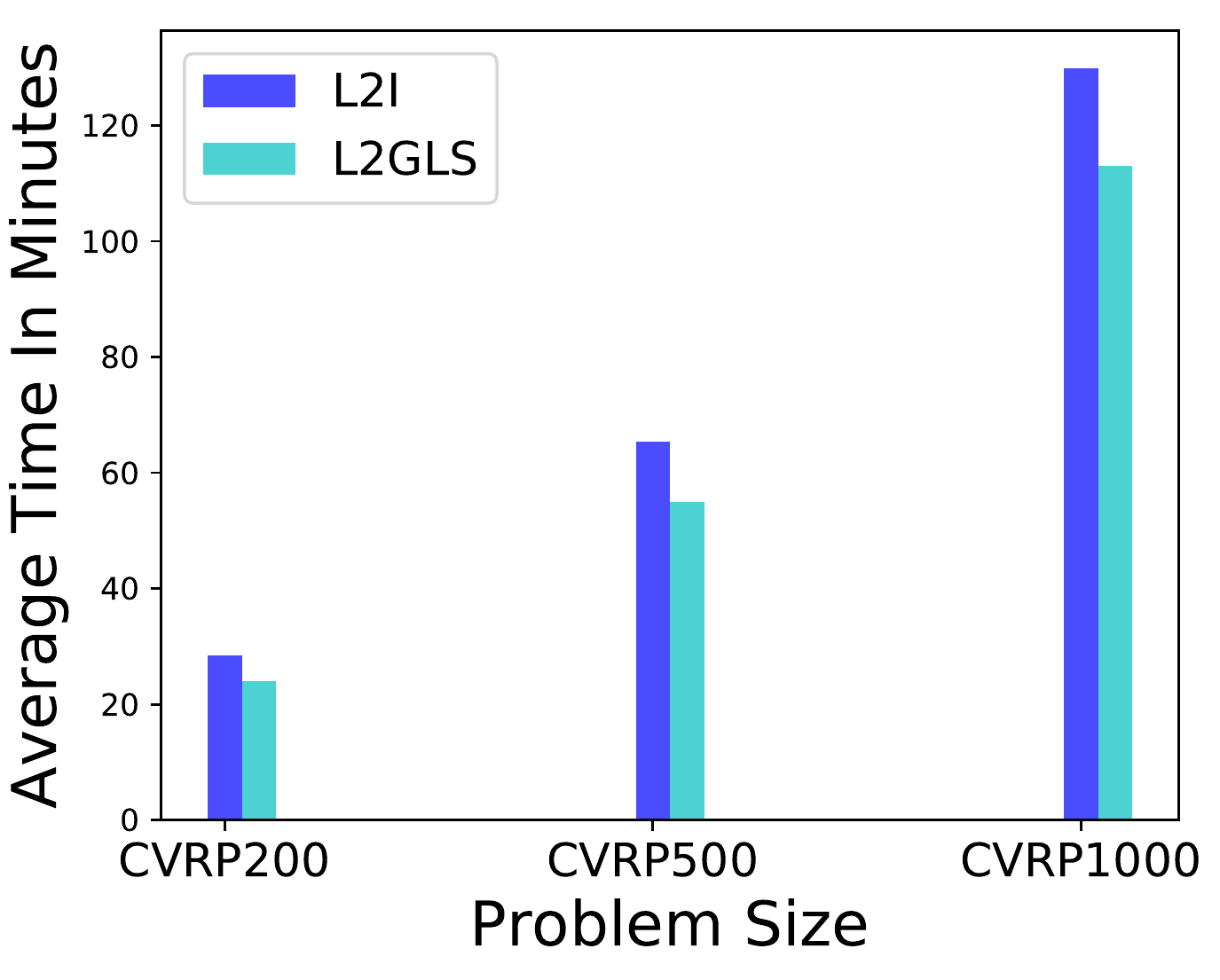}
    \caption{CVRP computation time.}\label{fig:vrpltime}
  \end{subfigure}
   \caption{Average computation time for larger TSP and CVRP.}\label{fig:timevrp}
\end{figure}
\begin{table*}[ht]
\centering
\caption{TSPLIB results: Instances are sorted by increasing size, with the number at the end of an instance's name indicating its size. Values reported are the cost of the tour found by each method (lower is better, best in bold). Gap (\%) is the gap to the solution obtained by Concorde~\cite{applegate2006concorde}. The results marked * reported from S2VDQN~\cite{khalil2017learning} and marked $\dagger$ reported from LHI~\cite{wu2019learning}. $**$ means (column: Gap\%) our model performed better than solver, no gap reported.}
\scriptsize
\begin{tabular}{|l|ll|ll|ll|ll|ll|ll|}\hline
Problems & \multicolumn{2}{|c|}{L2I} & \multicolumn{2}{|c|}{Or-tools}& \multicolumn{2}{|c|}{S2VDQN*} & \multicolumn{2}{|c|}{AM$\dagger$}& \multicolumn{2}{|c|}{LHI$\dagger$}& \multicolumn{2}{|c|}{L2GLS}\\ \hline
\textbf{}    & TourL  & Gap(\%)& TourL & Gap(\%)  & TourL & Gap(\%)  & TourL  & Gap(\%)& TourL & Gap(\%)  & TourL & Gap(\%) \\\hline 
Eil51 & 430 & 0.93& 436& 2.34  & 439 &3.05& 435&2.11& 438 &2.81& \textbf{428} & 0.46\\\hline
Berlin52& 7,694& 2.01 & 7,945& 5.34 & \textbf{7,542} &0.00 &  7,668 & 1.67& 8,020&  6.33&7,544& 0.026\\\hline
St70& 683& 1.85 & 683& 1.18  & 696& 3.11 &  690& 2.22 & 706& 4.59& \textbf{679} & 0.29\\\hline
Eil76& 551 & 2.41 & 561&  4.27 & 564& 4.83 & 563& 4.64 &575&6.87&  \textbf{546}& 1.48\\\hline
Pr76& 108,871& 0.65 & 111,104&  2.72 & {108,446}& 0.26 &  111,250&2.85  &109,668& 1.39& \textbf{108,159}&0.00 \\\hline
rat99& 1,257&3.79 & 1,232 &1.73 &  1,280& 5.69 &  1,319& 8.91 & 1,419&17.17  &\textbf{1,221}& 0.82\\\hline
KroA100& 22,036&3.54 & \textbf{21,448}& 0.78 &  21,897& 2.88 & 38,200& 79.49 & 25,196&  18.39& {21,650}&1.72 \\\hline
KroB100& 22,452& 1.40& 23,006&  3.90 &22,692& 2.48 & 35,511& 60.38 & 26,563& 19.97 &\textbf{22,191}& 0.22 \\\hline
KroC100&21,258 &2.45 &  21,583& 4.01&  21,074& 1.56& 30,642 & 47.67&25,343&   22.14& \textbf{20,820}&0.34 \\\hline
KroD100& 22,102& 3.79 & 21,636& 1.60  & 22,102& 3.79& 32,211 & 51.26&24,771&  16.32 &\textbf{21,514}&  1.03\\\hline
KroE100&22,875 &3.65 &22,598&   2.40&22,913 &3.82 &27,164 &  23.09& 26,903 & 21.90& \textbf{22,152}&4.02 \\\hline
rd100& 8,132& 2.80  & 8,189& 3.52&  8,159 &3.14 &8,152 & 3.05 &\textbf{7,915}& 0.06 & 7944& 0.42\\\hline
eil101& 654& 3.97& 664& 5.56 & 659 &4.76 &667 &6.04 &658&4.61 &\textbf{642}& 2.06 \\\hline
Lin105& 14,700& 2.23 & 14,824&  3.09&  15,023& 4.47 & 51,325&256.94&  18,194&26.53 &\textbf{14,406}& 0.18 \\\hline
pr107& 44,807&1.13 & 45,072& 1.73 & 45,113&  1.82 & 205,519&  363.89& 53,056& 19.75 & \textbf{44,301}& **\\\hline
pr124&59,935 & 1.53& 62,519& 5.91   & 61,623 &  4.39& 167,494& 183.74 & 66,010& 11.82 &\textbf{59,030}& 0.00 \\\hline
bier127& 120,699&2.04 & 122,733&  3.76 & 121,576&2.78 & 207,600& 75.51& 142,707& 20.64 &\textbf{119,281}& 0.84\\\hline
Ch130& {6,318}& 3.40 & 6,470&5.89  &6,270& 2.61& 6,316 &3.37 & 7,120& 16.53& \textbf{6.178}& 1.11\\\hline
pr136&  98,728& 2.02&102,213& 5.62  &  99,474& 2.79 & 102,877& 6.36& 105,618&9.14&  \textbf{98680}&0.50 \\\hline
pr144& 59,895&2.31 & 59,286&  1.27&  59,436& 1.53 & 183,583&  213.61& 71,006& 21.30 & \textbf{58,535}& ** \\\hline
Ch150 & 6,779& 3.84  & 7,232& 10.78  & 6,985 &7.00 & 6,877& 5.34& 7,916&21.26 &\textbf{6,606}& 1.19 \\\hline
KroA150& 27,307& 2.95&  27,592&4.02 &  27,888&5.14 & 42,335 & 59.61& 31,244& 17.79 &\textbf{27,248}& 2.72\\\hline
KroB150& 26,563&19.97 & 23,006& 3.90 & 27,209& 22.88 &35,511 & 60.38 &26,563& 19.97  &\textbf{22,381} & 1.08\\\hline
pr152& 75,196 &1.05& 75,834 & 2.92& 75,283 &2.17 &103,110 & 39.93& 85,616&16.19 & \textbf{74,204} &0.70 \\\hline
u159& 43,220 &2.70 & 45,778 & 8.78& 45433& 7.96 & 115,372&174.17 &51,327& 21.97& \textbf{42,075} & **\\\hline
rat195& 2,403 &3.44 & 2,389& 2.89 & 2,581&  11.10 &3,661& 57.59& 2,913& 25.39& \textbf{2,368}&  1.93\\\hline
d198& 16,349 &  3.60& 15,963&1.15&   16,453& 4.26 & 68,104&331.58 &17,962& 13.82 & \textbf{16,244} & 2.94\\\hline
KroA200& 30,617& 4.25& 29,741 & 1.27&  30,965 & 5.43&  58,643 &99.68 & 35,958 &22.43 &\textbf{29,736}&1.25 \\\hline
KroB200& 30,925&5.05&  30,516& 3.66& 31,692& 7.66 &50,867  &72.79 & 36,412& 23.69 & \textbf{30,154}& 2.43\\\hline
ts225& 129,638& 2.36 & 128,564&  1.51& 136,302& 7.62  &  141,628&  11.83& 158,748 &25.35 &\textbf{126,645} & 0.001\\\hline
Tsp225& 3,995 &2.01 & 4,046 &3.31 &  4,154 & 6.07&24,816&  533.7 &4,701 & 20.04 & \textbf{3,940}& 0.16 \\\hline
pr226& 87,160&8.44 & 82,968 &3.23&   81,873 & 1.87& 101,992 &26.90 & 97,348&21.12& \textbf{81,756}& 1.72 \\\hline
gil262& 2,534 & 6.56 & 2,519 &  5.92& 2,537 & 6.68& 2,693 &13.24 & 2,963 & 24.60&\textbf{2,451}&  3.06\\\hline
pr264& 56,191 & 14.36 &51,954& 5.73  & 52,364&  6.57& 338,506 & 588.93& 65,946 & 34.21& \textbf{50,528}& 0.00\\\hline
A280& 2,751&  6.66&  2713&5.19 &  2,867& 11.16 & 11,810 & 357.92& 2,989 &15.89 & \textbf{2,672}& 3.60\\\hline
Pr299& 53,753 &11.54 & 49,447&  2.60& \textbf{51,895} & 4.95& 513,673 & 938.83& 59,786 &20.90 & 52,827&  7.68\\\hline
Lin318& 44,151 & 5.04& -  & & 45,375 & 7.96& - &-& -& -&  \textbf{43,184}& 2.74\\\hline
Pr439& 113,124 & 5.50 & - & - &- &- &- & -& -& - & \textbf{111,269}& 3.77 \\\hline
Pr1002& 273,970 & 5.76  & - & - &- & -&- & -&- & - &  \textbf{262,150}& 1.19\\\hline
\end{tabular}
\label{tabile:tsplib}
\end{table*}

\subsection{Computational Running Time Analysis}\label{section:comtime}
Comparing the algorithms' efficiency in terms of execution time is difficult due to varying hardware and implementations among different approaches. So for fair comparison, we performed experiment on the same hardware we used for our work. We compare the inference time with a recant improvement based approach  L2I~\cite{lu2019learning} using their publicly available code~\footnote{https://github.com/rlopt/l2i}. For TSP  Figures~\ref{fig:tspstime}~\ref{fig:tspltime} illustrate that L2GLS is efficient compared to Concorde~\cite{applegate2006traveling}. For example, to solve TSP100, Concorde~\cite{applegate2006traveling} needs approximately 60 minutes, whereas L2GLS required only 11 mins to solve (Figure~\ref{fig:tspstime}). Given 1000 random instances. L2GLS significantly reduces the running time for all instances compare to Concorde~\cite{applegate2006traveling}~L2I~\cite{lu2019learning} for all problem sizes of TSP without sacrificing finding optimal tours, as shown in Figure~\ref{fig:tspstime}~\ref{fig:tspltime}. 
For CVRP, the running time for LKH3 is rapidly increases when the problem size is large (Figure~\ref{fig:vrpstime}). For example, to solve CVRP100, LKH3 needs 780 minutes, whereas L2GLS requires only 17 minutes, as shown in Figure~\ref{fig:vrpstime}. L2GLS allows notable reduction in the run time with good quality solution compare to L2I~\cite{lu2019learning}.

We also run experiments with larger instances with sizes up to 1000 nodes where results are summarised in Figure~\ref{fig:tspltime}. The Figure~\ref{fig:tspltime} shows that L2GLS saves more running time compared to Concorde~\cite{applegate2006concorde}. For example, in terms of TSP1000 problems, Concorde~\cite{applegate2006traveling} needs on average 796 minutes to solve a problem. However, L2GLS only need 47 minutes to solve the same problems, as shown in Figure~\ref{fig:tspltime}. Compared with L2I~\cite{lu2019learning}, L2GLS uses less average running time and tour length.

In the case of CVRP, It is evident from Figure~\ref{fig:vrpstime} that the running time of LKH3 grows with an increase in number of nodes, i.g., for CVRP100 size, LKH3 takes 780 minutes. Therefore, we fairly compared the running times with the L2I~\cite{lu2019learning} methods to show the efficiency of our method for CVRP200, CVRP500 and CVRP1000. The results are presented in Figure~\ref{fig:vrpltime}. Specifically, in terms of running time, our method not only is more efficient when compared to LKH3, but also outperformed L2I~\cite{lu2019learning}, as illustrated in Figure~\ref{fig:timetsp} and~\ref{fig:timevrp}.

It is worth noting that for smaller and larger randomly generated instances, L2GLS can generate solutions faster than the state of the art algorithms, i.e., Concorde for TSP and LKH3 for CVRP (\ref{fig:timetsp} and~\ref{fig:timevrp}). 

\begin{table}[ht]
\scriptsize
\centering
\caption{CVRP instances analysis using nine different scenarios, namely random, central and eccentric depot positioning combine with random, clustered and random-clustered customer positioning. Values reported are the cost of the tour found.}
\begin{tabular}{|c|c|c|c|c|c|c}\hline
 Problems & $=>$ & CVRP20 & CVRP50& CVRP100 \\\hline
\textbf{Depot Position} & \textbf{Customer Position}  &  & &  \\\hline
Random&  Random &5.86 & 10.37  &15.60  \\\hline
Random& Clustered &4.82  & 8.46 & 13.45 \\\hline
Random& Random-Clustered  &5.86  & 10.31 & 16.59\\\hline
Central& Random & 5.07 &8.79  & 13.59 \\\hline
Central& Clustered &  3.74& 6.43& 9.96 \\\hline
Central& Random-Clustered & 5.07 & 8.25 & 12.4  \\\hline
Eccentric& Random & 7.42 & 13.53 &  20.89 \\\hline
Eccentric& Clustered& 6.05& 10.86 & 17.46\\\hline
Eccentric& Random-Clustered & 7.35& 13.14 &  19.55 \\\hline
\end{tabular}
\label{tabile:cvrplib}
\end{table}
\subsection{Generalisation - L2GLS method}
In this section, we will show how our method can generalise well to benchmark libraries. Specifically we utilise TSPLIB instances~\cite{reinelt1991tsplib} which are generally used to compare TSP algorithms and data proposed by Uchoa et al.~\cite{uchoa2017new} for CVRP. In addition, we evaluate our method's generalisation performance on different problem sizes of TSP and CVRP, where we trained and tested on different size of benchmark datasets.
\subsubsection{Benchmark (TSPLIB)}
L2GLS model was trained on randomly generated 50 nodes of instances. Our aim in these experiments are to know to what extent the learned algorithm generalises to benchmark instances. The distribution of the instance, such as node locations, is completely different from those we used in the training instance. We compare our method with S2VDQN~\cite{khalil2017learning}, AM~\cite{kool2018attention}, LHI~\cite{wu2019learning} and L2I~\cite{lu2019learning}, the results are shown in Table~\ref{tabile:tsplib}, which demonstrates that L2GLS outperformed all the learning-based approaches most of the instances. Moreover, the heuristic solver OR-Tools generates inferior solutions compared to L2GLS~\cite{reinelt1991tsplib}.
\begin{table*}[ht]
\scriptsize
\centering
\caption{Comparison of our different methods (combination of LS operators), L2GLS, L2GLS2,L2GLS3, L2GLS without penalty term.}
     \begin{subtable}{0.45\linewidth}
    \begin{minipage}{0.45\linewidth}
      \caption{TSP results.}
\begin{tabular}{|l|l|l|l|}\hline
  & TourL   & TourL   & TourL \\\hline 
\textbf{TSP} & \multicolumn{1}{|c|}{TSP20} & \multicolumn{1}{|c|}{TSP50}& \multicolumn{1}{|c|}{TSP100} \\ \hline
L2GLS& 3.72  & 5.67   & 7.69 \\\hline
L2GLS2&  3.80  & 5.72   & 7.80 \\\hline
L2GLS3&  3.80  & 5.74   & 7.80 \\\hline
L2GLS without penalty&  3.72  & 5.69   & 7.72 \\\hline
        \end{tabular}
        \label{tabile:allpist}
    \end{minipage}%
    \end{subtable}%
     \begin{subtable}{.45\linewidth}
    \begin{minipage}{.45\linewidth}
      \centering
        \caption{CVRP results.}
        \begin{tabular}{|l|l|l|l|}\hline
          & TourL    & TourL   & TourL  \\\hline 
\textbf{CVRP} & \multicolumn{1}{|c|}{CVRP20} & \multicolumn{1}{|c|}{CVRP50}& \multicolumn{1}{|c|}{CVRP100} \\ \hline
L2GLS& 5.85  & 10.30   & 14.67 \\\hline
L2GLS2&  6.12  & 10.5 0  & 16.54 \\\hline
L2GLS3&  6.17  & 11.00   & 17.70 \\\hline
L2GLS without penalty&  5.86  & 10.37  & 15.60 \\\hline
        \end{tabular}
        \label{tabile:allpisv}
    \end{minipage} 
    \end{subtable}%
\end{table*}
We tested L2GLS methods on the 39 TSP instances up to size 1002 instance from TSPLIB~\cite{reinelt1991tsplib} in Table~\ref{tabile:tsplib}. For all the scenarios, L2GLS obtain a small gap compare to Concorde~\cite{applegate2006concorde} and outperformed all the neural network-based approaches. All the recent works only evaluated their methods up to 299 nodes, but to show the capability of our approach, we tested another three instances, namely Lin318, Pr439 and Pr1002. As shows in Table~\ref{tabile:tsplib}, our algorithm outperforms the prior ML based approaches in terms of solution quality. For three large instances we compare our method with L2I~\cite{lu2019learning}. Our L2GLS optimality gap is 2.74\% 3.77\% and 1.19\%, whereas L2I~\cite{lu2019learning} gap is 5.04\% 5.50\% and 5.76\% for Lin318, Pr439 and Pr1002 problem instances respectively. Table~\ref{tabile:tsplib} also shows L2GLS outperformed Concorde~\cite{applegate2006concorde} for some instances (denoted as $**$ in Table~\ref{tabile:tsplib}), such as pr107, pr144 and u159. However, we emphasise that the aim is not to outperform Concord but to show L2GLS performs well. 
\subsubsection{Practical Instances (CVRP)}\label{section:benchmark}
This section evaluated L2GLS methods with the CVRP benchmarks~\cite{uchoa2017new}, where each instance is characterised by the following attributes: number of customers, depot positioning, and customer positioning. The details of each attribute are described in the supplementary. This data distribution is commonly used in the OR field. 
We have evaluated L2GLS on various scenarios of CVRP benchmark instances. We utilised three sizes of the CVRP dataset with some additional scenarios following the data generation proposed in~\cite{uchoa2017new}. We further combined the customer and depot positioning and reported nine scenarios, namely random, central and eccentric depot positioning, combined with random, clustered and random-clustered customer positioning. Table~\ref{tabile:cvrplib} shows the impact of depot positioning while using different customer position. This experiment which demonstrates that L2GLS works over different data distribution ( on nine different data distributions).

\begin{table}[ht]
\scriptsize
\centering
\caption{Compared two L2GLS variants: L2GLS and L2GLS without penalty term, results are reported for TSPLIB. $**$ means (column: Gap\%) our model performed better than solver, no gap reported. Bold means better.}.

\begin{tabular}{|l|ll|ll|ll|}\hline
Problems & \multicolumn{2}{|c|}{Concorde} & \multicolumn{2}{|c|}{L2GLS without Penalty}& \multicolumn{2}{|c|}{L2GLS}\\ \hline
\textbf{}    & TourL  & Gap(\%)& TourL & Gap(\%)   & TourL & Gap(\%) \\\hline 
St70& 675& 0.00  & 679&  0.59& \textbf{677} & 0.29\\\hline
Pr76& 108,159& 0.00  &108,570& 0.37& \textbf{108,159}&0.00 \\\hline
rat99& 1,211&0.00   & 1,223&0.99  &\textbf{1,221}& 0.82\\\hline
KroA100& 21,282&0.00 & 21,866& 2.74 & \textbf{21,650}&1.72 \\\hline
rd100& 7,910& 0.00 &8,010& 1.26 & \textbf{7944}& 0.42\\\hline
Lin105& 14,379& 0.00 &  14,565& 1.29&\textbf{14,406}& 0.18 \\\hline
pr107& 44,303&0.00 & 44,527& 0.50 & \textbf{44,301}& **\\\hline
Ch130& {6,110}& 0.00 & 6,192& 1.39& \textbf{6,178}& 1.11\\\hline
pr144& 58,537&0.00 & 58,781& 0.44 & \textbf{58,535}& **\\\hline
KroA150& 26,524& 0.00& 27,408& 3.33 &\textbf{27,248}& 2.72\\\hline
u159& 42,080 &0.00 &42,645& 1.35 & \textbf{42,075} & **\\\hline
d198& 15,780 &  0.00 &16,341& 3.55 & \textbf{16,244} & 2.94\\\hline
ts225& 126,643& 0.00  & 129,248 & 2.05&\textbf{126,645} & 0.001\\\hline
Tsp225& 3,916 &0.00  &3,940 & 2.83 & \textbf{3,940}& 0.16 \\\hline
Pr299& 48,191 &0.00 & 52,827 &9.62 &51,895 &  7.68\\\hline
\end{tabular}
\label{tabile:abla}
\end{table}
\subsection{Ablations}\label{section:impact}
To evaluate the value of a different set of LS operators, we run four variants of L2GLS, i.e., L2GLS, L2GLS2, L2GLS3 and L2GLS without penalty terms. \textbf{L2GLS} consists of $(2opt+relocate+swap+three permutation)$ with penalty term that our main method, \textbf{L2GLS2} consists of $(2opt+relocate+swap)$, \textbf{L2GLS3} consists of $(2opt+swap+three permutation)$, and \textbf{L2GLS without penalty terms} consists of operators $(2opt+relocate+swap+three permutation)$ without the penalty term.

The goal of this experiment is to analyse the impact of a different set of LS operators. As we mentioned earlier we implemented and experimented with a set of LS operators and concluded with the L2GLS method, which is a combination of 2-opt, relocate, swap, and three permutations with penalty terms that is computationally efficient and able to produce state-of-the-art results for TSP.
\subsubsection{Evaluation of the four variants of L2GLS on Random TSP and CVRP instances}
Tables~\ref{tabile:allpist} and~\ref{tabile:allpisv} illustrate the performance of a different set of search operators to solve random TSP and CVRP problem instances. Each variant used the same neural network architecture, to evaluate the effect of the search operators on the solving results. It can be observed that using a different variant of operators impacts the solution quality. The best TSP and CVRP results are obtained by L2GLS among all the variants. From Tables~\ref{tabile:allpist} and~\ref{tabile:allpisv}, it can be seen that L2GLS3, by not using relocate, generates significantly worsen results for TSP and CVRP.

\subsubsection{Evaluation of two variants of L2GLS on TSPLIB instances}
Among the four variants, L2GLS and L2GLS without the penalty term performed close to the optimal solution Concorde~\cite{applegate2006concorde} shown in Tables~\ref{tabile:allpist} and~\ref{tabile:allpisv} for random instances. Therefore, we selected benchmark from our Table~\ref{tabile:tsplib} and evaluated our model without penalty term to check how our model performed. From Table~\ref{tabile:abla}, it is clear that using penalty term with LS operators significantly improves over only using LS operators to solve the problems. 

We demonstrated that our method outperformed recent approaches. The potential for better performance is the choice of heuristics used as LS operators. As many LS operators are effective and efficient for solving COP, therefore, in~\cite{arnold2019knowledge} states that well-implemented LS creation can compete with the best heuristics. Misztal et al.,~\cite{misztal2019impact} studied that swap belongs to a group of the LS algorithm characterised by the good quality of the returned solution. In~\cite{sengupta2019local} they concluded the 2-opt is the best individual operator, besides the mixed variant of operators found the optimum solution for many cases. In this work, we investigate different operators, which is a mixed variant of operators $(2opt+relocate+swap+three permutation)$. Moreover, we have introduced penalty terms, which contributes to significantly improved results for RPs. When searching stuck in local minima, we use penalty terms to improve the solution further; therefore, our result shows that these combined operators yield better solutions for most instances from TSPLIB~\cite{reinelt1991tsplib}. 

\begin{table}[ht]
\centering
\scriptsize
    \caption{Trained with TSP50. Average tour length, gap percentage (lower is better, best in bold).}
\begin{tabular}{|l|l|l|l|}\hline
  & TourL   & TourL   & TourL \\\hline 
     \textbf{TSP} & \multicolumn{1}{|c|}{TSP20} & \multicolumn{1}{|c|}{TSP50}& \multicolumn{1}{|c|}{TSP100} \\ \hline
Concorde&  3.84  & 5.70  & 7.77 \\\hline
L2GLS&  3.72  & 5.69   & 7.72 \\\hline
\textbf{CVRP} & \multicolumn{1}{|c|}{CVRP20} & \multicolumn{1}{|c|}{CVRP50}& \multicolumn{1}{|c|}{CVRP100} \\ \hline
LKH3~\cite{helsgaun2017extension}&  6.14  &   10.39 & 15.67 \\\hline
L2GLS & 5.86   &  10.52 & 15.9  \\\hline
        \end{tabular}
        \label{tabile:traintsp}
        \end{table}
    \begin{table}[h]
      \centering
      \scriptsize
        \caption{Trained with CVRP50.Average tour length, gap percentage (lower is better, best in bold)}
        \begin{tabular}{|l|l|l|l|}\hline
          & TourL    & TourL   & TourL  \\\hline 
     \textbf{TSP} & \multicolumn{1}{|c|}{TSP20} & \multicolumn{1}{|c|}{TSP50}& \multicolumn{1}{|c|}{TSP100} \\ \hline
Concorde&  3.84  & 5.70  & 7.77 \\\hline      
L2GLS & 3.76   & 5.70 & 7.76 \\\hline
\textbf{CVRP} & \multicolumn{1}{|c|}{CVRP20} & \multicolumn{1}{|c|}{CVRP50}& \multicolumn{1}{|c|}{CVRP100} \\ \hline

LKH3&  6.14  &   10.39 & 15.67 \\\hline
L2GLS & 5.85    & 10.30  & 14.67  \\\hline
        \end{tabular}
        \label{tabile:vrptrain}
\end{table}

\subsection{Various test and training}\label{appendix:generalisation}
One of the aims of the proposed methods was to ensure its generalisation on different problem sizes of TSP and CVRP. We trained the model with TSP50, used to solve TSP20, TSP50, TSP100, and CVRP20, CVRP50, CVRP100 presents the result in Table\ref{tabile:traintsp}. We trained the same model with TSP50, used to solve TSP20, TSP50, TSP100, CVRP20, CVRP50, CVRP100, and present the result in Table\ref{tabile:vrptrain}. We show that L2GLS can generalise to different problem distributions even when trained on different sizes and problems. 
\section{Conclusion}
We proposed L2GLS for solving RPs, which starts with an initial solution and improves the solution with LS operators selected by an RL-based manager or with penalty terms to guide the LS further to improve the solution. We also analysed distinct choices of LS operators and selected the best choice of operators from this analysis. The current need is to find an effective combination of LS operators that can generate optimum solutions that adapt and generalise to different problems, which our reinforcement-based manager achieved.  
Among the four design choices discussed in Section~\ref{section:impact}, we selected L2GLS because we achieve better solution quality applied L2GLS on TSP and relative better for CVRP. Many learning algorithms can solve small, randomly generated problems. However, many applications in the real world are related to larger instances; there is also variability in data distribution. Most of the state of the art algorithms have challenges to solve these practical problems. Therefore, we proposed a method to choose effective LS operators and let RL-based managers learn all the problems to predict various other RPs. L2GLS achieved new state-of-the-art results for TSP instances. Moreover, it outperforms not only for small scale problems but can generalise well benchmark and large-scale datasets. For CVRP, it outperforms LKH3, OR-tools~\cite{perron2019google} and recent DL-based baselines. 

Future research can explore expanding our solution framework to solve other variants of combinatorial problems. Moreover, advanced search operators can be utilised for other combinatorial problems.

\bibliographystyle{unsrt}
\bibliography{my}

\newpage

\appendix

\begin{figure}[!h]
  \vspace{-0.2cm}
  \centering
  \includegraphics[width=0.4\textwidth]{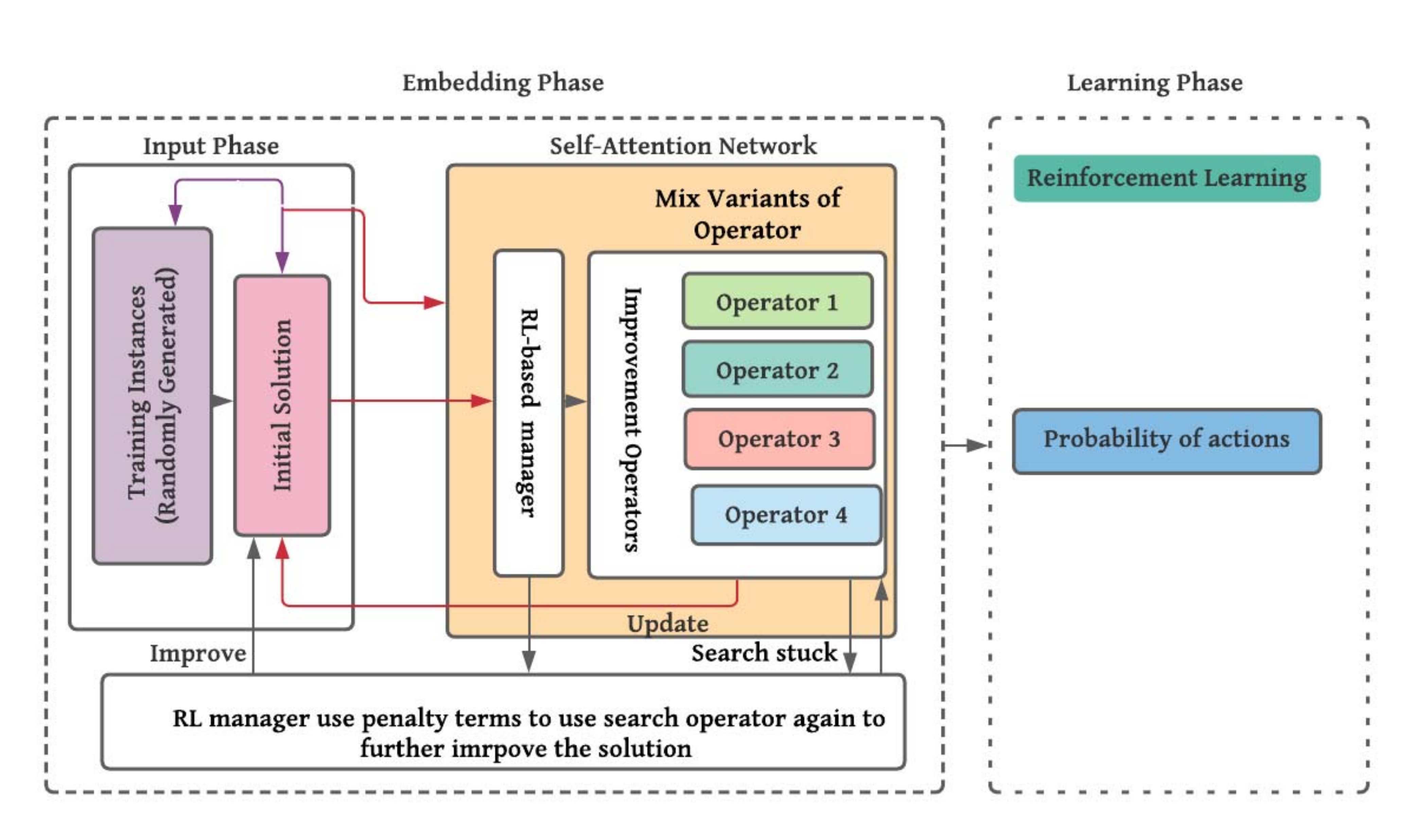}
  \caption{Overview of policy network. The first box is the state embedding part of policy network. Embedding part contains problem features and solution features, an attention network, and a sequence of actions and their effects.}\label{fig:policy}
  \vspace{-0.1cm}
\end{figure}

\subsection{Policy Network}\label{appendix:reward}
In our policy network problem and solution-specific input features are transformed into an embedding, which is fed into an attention network. The output of the attention network is concatenated with a sequence of recent actions and their effects. In the end, the output of the attention network is fed into two fully connected layers. Figure~\ref{fig:policy} shows our overall policy network.

\subsection{Instance attributes: CVRP}\label{appendix:cdpo}
Instance attributes, depot positioning Three different positions for the depot are considered: Central (C) – depot in the centre of the grid, Eccentric (E) – depot in the corner of the grid, point and Random (R) – depot in a random point of the grid. Customer positioning Three alternatives for customer positioning are considered: following the R, C and RC instance classes of the Solomon set for the VRPTW~\cite{solomon1987algorithms}. Random (R) – all customers are positioned at random points of the grid. Clustered (C) –  A number N of customers that will act as cluster seeds are picked from a uniform discrete distribution. Next, the N nodes are randomly positioned in the grid following~\cite{uchoa2017new}. Random-clustered (RC)– Some customers are clustered using Cluster positioning; the remaining customers are randomly positioned.

\end{document}